\documentclass[lettersize,journal]{IEEEtran}
\usepackage{amsmath,amsfonts}
\usepackage{algorithmic}
\usepackage{algorithm}
\usepackage{array}
\usepackage[caption=false,font=normalsize,labelfont=sf,textfont=sf]{subfig}
\usepackage{textcomp}
\usepackage{stfloats}
\usepackage{url}
\usepackage{verbatim}
\usepackage{graphicx}
\usepackage{cite}
\usepackage{multirow}
\usepackage{float}
\usepackage{graphicx}
\usepackage{caption}
\usepackage{subcaption}
\usepackage{pifont}

\usepackage{colortbl}
\usepackage[table]{xcolor}
\usepackage{booktabs}
\usepackage[margin=0.75in]{geometry}
\usepackage{booktabs}
\usepackage{multirow}
\usepackage{amsmath,amssymb}
\usepackage{xcolor}
\usepackage{colortbl}
\usepackage{makecell}
\usepackage{graphicx}
\usepackage{subcaption}
\usepackage{pifont}
\usepackage{adjustbox}
\usepackage[caption=false,font=normalsize,labelfont=sf,textfont=sf]{subfig}
\usepackage{bm}

\definecolor{lightgray}{gray}{0.93}
\newcommand{\ours}{\rowcolor{lightgray}}

\def\BibTeX{{\rm B\kern-.05em{\sc i\kern-.025em b}\kern-.08em
    T\kern-.1667em\lower.7ex\hbox{E}\kern-.125emX}}
\usepackage{balance}
\begin{document}
\title{HACK++: Towards More Effective Head-Aware Key-Value Compression for Efficient Visual Autoregressive Modeling}
\author{Ziran~Qin,
        Yuchen~Jiang,
        Mingbao~Lin$^{\dagger}$,
        Youru~Lv,
        Hang~Guo,\\
        Fei~Wen,\,\textit{Senior Member,\,IEEE},
        Weiyao~Lin$^{\dagger}$,\,\textit{Senior Member,\,IEEE}
\thanks{Ziran Qin, Yuchen Jiang, Youru Lv, Fei Wen, and Weiyao Lin are with Shanghai Jiao Tong University, Shanghai, China; Mingbao Lin is with Rakuten, Singapore. Hang Guo is with Tsinghua University, Beijing, China. E-mail:\{qinziran, wylin\}@sjtu.edu.cn.}
\thanks{$^{\dagger}$Corresponding authors: Mingbao Lin and Weiyao Lin.}}

\markboth{Journal of \LaTeX\ Class Files,~Vol.~18, No.~9, September~2020}%
{How to Use the IEEEtran \LaTeX \ Templates}

\maketitle

\begin{abstract}
Visual Autoregressive (VAR) models adopt a next-scale prediction
paradigm, offering high-quality content generation with substantially
fewer decoding steps. However, existing VAR models suffer from
significant attention complexity and severe memory overhead due to
the accumulation of key--value (KV) caches across scales. In this
paper, we tackle this challenge by introducing KV cache compression
into the next-scale paradigm. We begin with an in-depth
analysis of VAR attention and observe that attention heads can be
stably divided into two functionally distinct categories:
\textit{Contextual Heads} focus on maintaining semantic consistency,
while \textit{Structural Heads} are responsible for preserving
spatial coherence. Their geometrical and functional divergence makes
existing one-size-fits-all compression methods perform poorly on VAR
models. Building on this divergence, we further find that the two
head types differ markedly in their reliance on historical scales,
and that this reliance also shifts across layers and generation
steps, arguing for an adaptive cache budget allocation. To address
these challenges, we propose \textbf{HACK++}, a training-free
\textbf{H}ead-\textbf{A}ware key-value \textbf{C}ompression
framewor\textbf{K} for VAR models. From a one-time offline
calibration, HACK++ classifies head types and derives head-specific
priors. At inference, it decouples attention from cache compression
under independent budgets, bounding the current-scale attention cost
while compressing the accumulated cache far more aggressively, via
pattern-specific strategies and a reliance-aware budget allocation. Extensive experiments on multiple VAR models
across text-to-image, class-conditional, and unified
understanding-and-generation tasks validate the effectiveness and
generalizability of HACK++. By aggressively reducing both attention computation and KV-cache size without degrading output quality, HACK++ delivers substantial memory savings and faster inference. For example, on Infinity-2B/8B,
HACK++ maintains near-lossless generation with only a $30\%$
attention budget and a $10\%$ cache budget, and remains robust
even under an extreme $1\%$ cache budget.
\end{abstract}

\begin{IEEEkeywords}
Visual Autoregressive Modeling, Image Generation, Efficient Inference.
\end{IEEEkeywords}

\section{Introduction}
\label{sec:intro}

\IEEEPARstart{A}{utoregressive} (AR)
models~\cite{he2024mars, sun2024autoregressive, li2024autoregressive,chen2025janus,wu2024janus,chameleonteam2025chameleonmixedmodalearlyfusionfoundation}
have shown promising performance in visual generation, rivaling diffusion models~\cite{ho2020denoising, podell2023sdxl,
peebles2023scalable, esser2024scaling, chen2024pixart} in both quality
and flexibility. Conventional AR models, however, follow a next-token
prediction paradigm whose long decoding chains incur substantial
inference latency. Visual AutoRegressive (VAR)
modeling~\cite{tian2024visual} addresses this by
reframing image synthesis as \emph{next-scale} prediction, generating
an image as a coarse-to-fine sequence of multi-scale token maps. This
design retains the sequential modeling capacity of autoregressive
methods while enabling highly parallel decoding within each scale,
yielding competitive generation quality in only a few decoding steps.
Consequently, VAR has been rapidly adopted across a wide
range of visual generation tasks and now serves as a backbone for many
state-of-the-art generative models.

\begin{figure*}[t]
  \centering
  \includegraphics[width=0.9\linewidth]{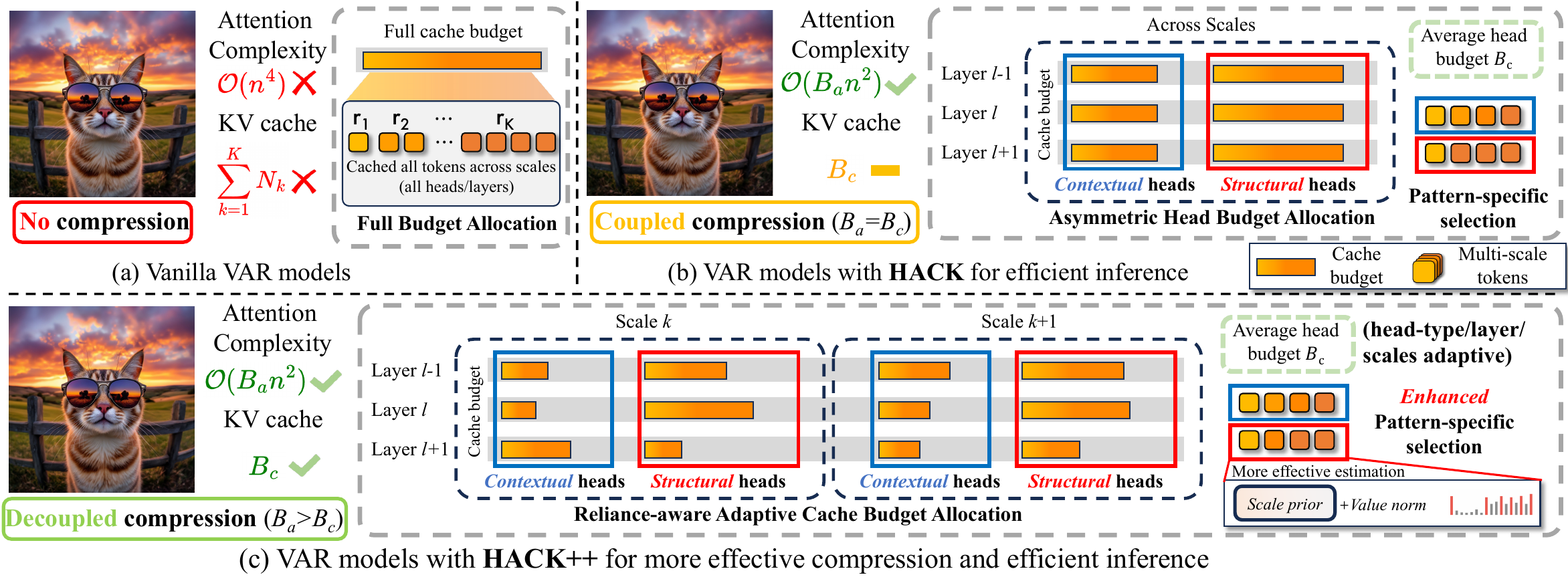}
\caption{(a) Vanilla VAR caches all KV pairs across scales without compression. (b) HACK applies a coupled compression pipeline ($B_a\!=\!B_c$) with head-aware, pattern-specific estimation and a fixed asymmetric allocation. (c) HACK++ hacks down both attention complexity and cache length via a decoupled pipeline ($B_a\!>\!B_c$) for more aggressive cache reduction, with enhanced pattern-specific estimation and a reliance-aware budget adapting across head types, layers, and scales.}
  \label{intro}
\end{figure*}

Despite these strengths, the next-scale formulation introduces a
severe inference bottleneck. As shown in Fig.\,\ref{intro}(a),
vanilla VAR indiscriminately caches the key-value (KV) pairs of all
historical scales, so the attended sequence at each step grows
cumulatively as generation proceeds, incurring even greater KV
cache accumulation than conventional AR models. This has two
compounding consequences: the per-step attention computation
becomes increasingly expensive as the attended sequence lengthens,
and the KV cache accumulates across scales until it dominates
memory consumption. The problem is especially acute for
high-resolution synthesis, where escalating attention cost and an
ever-growing KV cache make memory and latency the primary obstacles
to scale VAR models to higher resolutions and larger backbones.

Drawing inspiration from KV cache optimization in large language
models (LLMs), a natural remedy is KV cache compression, which
alleviates the attention and memory inflation caused by accumulated
KV states. However, existing LLM-oriented compression methods fail
to generalize to VAR models, due to fundamental differences in both
their generation paradigm and their attention mechanism. In terms
of paradigm, LLM inference separates a one-time prefilling
stage from sequential single-token decoding, and most compression
methods are tailored to one of these two stages; VAR instead
interleaves prefilling-and-decoding behaviors at every scale,
where each step both consumes the accumulated history and produces
a full new block of tokens. In terms of attention, the differences
in modality and task endow VAR with attention behaviors that differ
markedly from those of LLMs, yet remain largely unexplored.
Applying LLM compression rules that ignore these differences
indiscriminately discards tokens that VAR attention critically
depends on, causing severe quality degradation at high compression
ratios.

To tackle this, we propose \textbf{HACK++}, an efficient framework
that jointly compresses attention computation and KV cache in VAR
models. We first identify that the two sources of inefficiency, current-scale attention computation and the KV cache carried to future scales, serve different purposes and tolerate fundamentally different levels of compression: the former must retain enough context for accurate current-scale generation, whereas the latter, dominated by a few influential historical scales, admits far more aggressive reduction. Building on this, HACK++ introduces a \emph{decoupled} compression framework that separates attention compression from cache compression and governs each with its own budget. 
Then, through an in-depth analysis, we observe that VAR
attention heads are functionally heterogeneous and fall into two
distinct types: \emph{contextual} heads, which attend to a small
set of semantically salient tokens and anchor global semantic
content, and \emph{structural} heads, which attend in a
position-sensitive manner across scales and maintain spatial
structure. This functional specialization makes any uniform
compression rule mismatched with VAR attention, motivating a \emph{head-aware} strategy that selects tokens pattern-specifically according to each head type's role. 
Finally, we further observe that the reliance on historical scales is
itself highly uneven, varying across head types, layers, and generation steps, so a single uniform budget wastes capacity on caches that future steps rarely revisit. We therefore design an \emph{adaptive cache budget allocation} that distributes the limited cache budget to where it is most needed.

This paper is an extension of our conference work~\cite{qin2026head} where our
contributions were: (1)~an in-depth analysis that identifies and
characterizes contextual and structural heads in VAR models,
revealing their consistent functional roles and attention patterns;
and (2)~HACK, a head-aware compression framework that jointly
compresses attention and KV cache for VAR models through asymmetric
cache budget allocation and pattern-specific compression strategies
suited to both head types (summarized in Fig.\,\ref{intro}(b)).

In this journal version, illustrated in Fig.\,\ref{intro}(c), we
make several substantial
improvements: (1)~a more efficient pipeline that \emph{decouples}
attention and KV cache compression, enabling aggressive cache
reduction without degrading attention quality; (2)~an enhanced, head-aware importance estimation that more accurately identifies the critical tokens of each head type through pattern-specific criteria tailored to contextual and structural heads; (3)~a reliance-aware cache
budget allocation that adapts across head types, layers, and
generation steps; and (4)~an extensive evaluation on
text-to-image and class-conditional generation, together with an
extended study on unified VAR models, showing that HACK++ attains greater attention computation and KV cache reduction than existing KV cache compression pipelines while maintaining the best generation quality.

\begin{figure*}[!t]
  \centering
  \includegraphics[width=\linewidth]{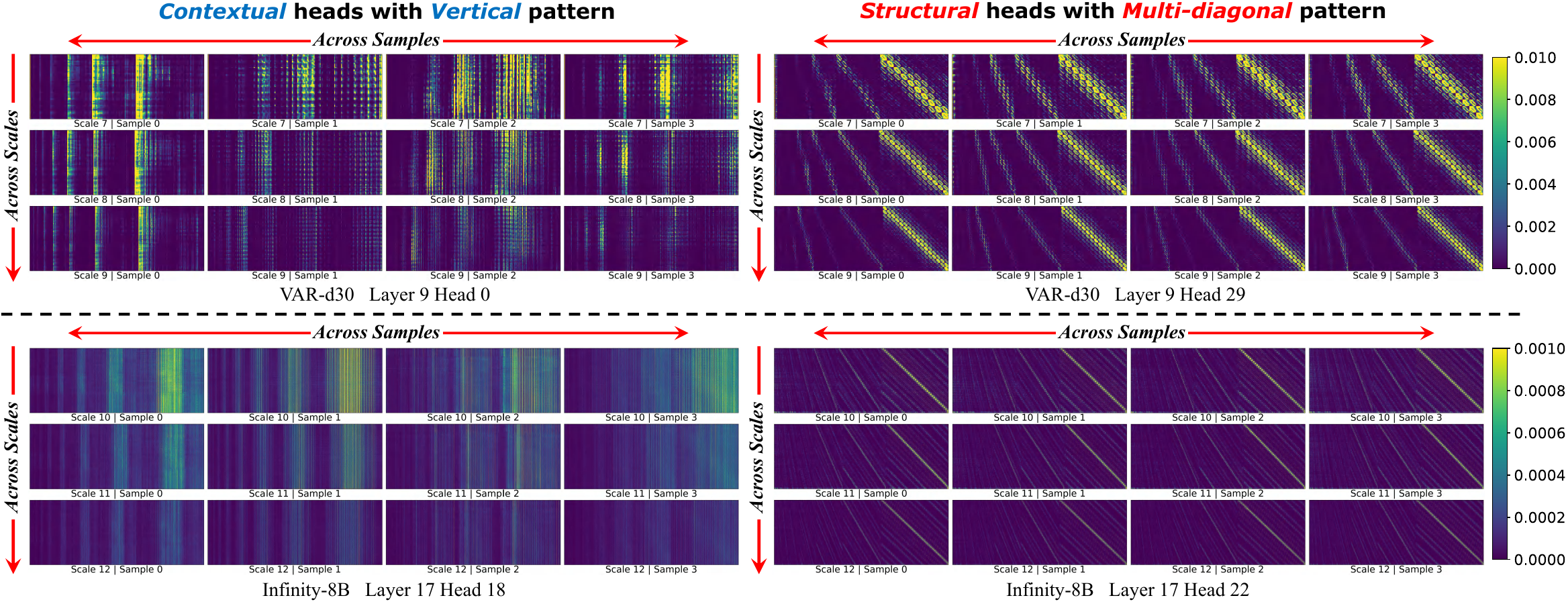}
  \caption{
    \textbf{Attention Patterns of Contextual and Structural Heads.}
    Both Contextual and Structural heads exhibit consistent vertical and multi-diagonal patterns, respectively, across different samples and scales.}
    \label{obser1}
\end{figure*}

\section{Related Work}
\noindent\textbf{Autoregressive Visual Generation.}
AR models~\cite{achiam2023gpt,anthropic2024claude3,dubey2024llama,liu2024deepseek,yang2025qwen3}, originally successful in LLMs, have been extended to the visual domain.
Traditional visual AR approaches~\cite{he2024mars,sun2024autoregressive,li2024autoregressive} rely on next-token prediction, where images are first quantized into discrete tokens (\emph{e.g.}, VQVAE~\cite{van2017neural}, VQGAN~\cite{esser2021taming}) and then decoded autoregressively. While effective, they incur high computation and quantization errors, resulting in lower efficiency.
Recent VAR models~\cite{tian2024visual,han2024infinity,tang2024hart,chen2024toward,ren2024m} instead adopt a next-scale prediction paradigm, generating an entire token map per step and enabling multi-scale parallel decoding that improves both quality and speed.
This paradigm has rapidly extended across diverse settings, including high-resolution text-to-image synthesis~\cite{han2024infinity,voronov2024switti}, video synthesis~\cite{liu2026infinitystar,ji2026videoar}, controllable generation~\cite{li2024controlvar}, image super-resolution~\cite{qu2025visual}, 3D content creation~\cite{chen2024sar3d,gao2025mars}, and unified understanding-and-generation models that handle both visual perception and synthesis within a single next-scale backbone~\cite{zhuang2025vargpt,Li2025OneCATDA}.
Despite this breadth, the hierarchical design causes attention complexity and KV cache to grow rapidly across scales, creating coupled memory and compute bottlenecks that hinder practical deployment.
We address these challenges via VAR-specific KV cache compression that reduces resource cost while preserving generation quality.

\noindent\textbf{KV Cache Compression.}
KV cache compression has been extensively studied to alleviate the
cache overhead of autoregressive models such as LLMs and VLMs.
Quantization reduces the cache footprint by lowering the numerical
precision of the stored key--value states~\cite{liu2024kivi,
yue2024wkvquant, kang2024gear, he2024zipcache}. Eviction and merging
strategies, in contrast, shorten the cache itself, either by
discarding less informative tokens~\cite{zhang2024h2o, li2024snapkv,
liu2024scissorhands, oren2024transformers, ren2024efficacy,qin2025autoregressive} or by
fusing redundant KV pairs~\cite{wan2024look, zhang2024cam,
liu2024minicache, wan2024d2o}. Our work follows this latter line,
curbing the excessive accumulation of cached tokens through eviction.
Most recent advances along this direction improve eviction and
merging by designing more principled retention criteria and budget
allocation schemes, typically grounded in an analysis of attention
behavior in LLMs and VLMs~\cite{xiao2023efficient,
xiao2025duoattention, qin2025cake, tu2025vl}. Effective as they are in those settings, however, these methods do not transfer readily to VAR
models, as the differences in generation paradigm, modality, and task
reshape the underlying attention dynamics.
Our method begins with an in-depth analysis of VAR attention, which in turn
informs a compression design tailored to VAR models---one that
delivers substantial memory and compute savings without compromising
visual quality.

\noindent\textbf{Efficient Autoregressive Visual Generation.}
Improving the inference efficiency of VAR has attracted growing interest, yet existing methods each leave a distinct limitation unaddressed.
Model-level approaches adjust computational capacity across scales: collaborative decoding~\cite{chen2024collaborative} schedules a large drafter and a small refiner over coarse and fine scales, while supernet-based methods~\cite{chen2025progressive} exploit the scale--depth dependency to assign shallower subnets to later scales within a single model.
Both, however, rely on auxiliary models or dedicated (re-)training and thus cannot serve as plug-and-play accelerators.
Token-pruning and sparsification methods~\cite{guo2025fastvar,chen2025frequency,chen2026toprovar} either prune tokens or skip certain scales and generation steps at later, high-resolution stages; focusing on computation reduction, their coarse-grained operations lack the means to maintain pixel-level fidelity.
Sparse-attention methods~\cite{xie2024litevar,li2026sparvar} cut attention cost by exploiting structured sparsity across scales, but typically depend on specialized attention kernels or hand-crafted sparse patterns, limiting their portability across architectures.
The above methods primarily reduce computation, yet offer little relief from the heavy memory footprint of the accumulated KV cache.

Closest to our work, our preliminary study HACK~\cite{qin2026head} first introduces KV cache compression to VAR, compressing the cache prior to attention to curb both attention computation and cache growth. The concurrent ScaleKV~\cite{li2025scalekv} instead enables an after-attention compression pipeline at layer granularity by distinguishing high- and low-demand layers, substantially reducing memory but offering limited control over the current-scale attention cost. Advancing this line, HACK++ decouples attention and cache compression under independent budgets, bounding both the current-scale attention cost and the long-term cache footprint. Beyond these distinctions, prior methods are each validated on a single task, leaving their generality across paradigms untested. HACK++ further builds on the contextual--structural head dichotomy, an intrinsic, task-invariant property of trained VAR attention rather than a task-specific heuristic, making it a general-purpose framework that we validate across class-conditional, text-to-image, and unified understanding-and-generation models. Notably, our method is orthogonal to quantization, token pruning, and VAR calibrated decoding, and can be combined for further efficiency gains.

\section{Preliminaries}

\textbf{Visual AutoRegressive Modeling.} VAR extends autoregressive modeling from ``next-token'' to ``next-scale'' prediction. Given an image feature map $\mathbf{f} \in \mathbb{R}^{h \times w \times c}$, VAR quantizes it into $K$ multi-scale token maps $\mathbf{R} = (\mathbf{r}_1, \mathbf{r}_2, \ldots, 
\mathbf{r}_K)$. Each map $\mathbf{r}_k \in [V]^{h_k \times w_k}$ contains $N_k=h_k \times w_k$ tokens and the resolutions $\{(h_k, w_k)\}_{k=1}^{K}$ are predefined such that $h_k w_k = a^{2(k-1)}$, where $a > 1$ is a constant scaling factor.
The autoregressive likelihood is:
\begin{equation}
    p(\mathbf{r}_1, \mathbf{r}_2, \ldots, \mathbf{r}_K) = \prod_{k=1}^{K} p(\mathbf{r}_k | \mathbf{r}_1, \mathbf{r}_2, \ldots, \mathbf{r}_{k-1}),
\end{equation}
where the generation of each scale $k$ is conditioned on the tokens $\{\mathbf{r}_1, \mathbf{r}_2, \ldots, \mathbf{r}_{k-1}\}$ from all preceding scales, which serve as a contextual prefix.
During inference,  the entire content is synthesized coarse-to-fine in just $K$ steps, with all $N_k$ tokens at each scale predicted in parallel. 

\textbf{Inefficiency in VAR.}
At each scale $k$, the input tokens $\mathbf{X}_k \in \mathbb{R}^{N_k \times D}$ are projected into queries, keys, and values:
\begin{equation}
\small
\mathbf{Q}_k = \mathbf{X}_k \mathbf{W}_Q, \quad
\mathbf{K}_k = \mathbf{X}_k \mathbf{W}_K, \quad
\mathbf{V}_k = \mathbf{X}_k \mathbf{W}_V,
\end{equation}
where $\mathbf{W}_Q, \mathbf{W}_K, \mathbf{W}_V \in \mathbb{R}^{D \times D}$ are the projection matrices, and $\mathbf{Q}_k, \mathbf{K}_k, \mathbf{V}_k \in \mathbb{R}^{N_k \times D}$ are the queries, keys, and values of the current scale. To avoid redundant computation, VAR caches KV pairs across scales. The current-scale keys and values are concatenated to the accumulated cache:
\begin{equation}
\small
\label{eq:kvcat}
\mathbf{K}_{\le k} = \text{Concat}(\mathbf{K}_{\le k-1}, \mathbf{K}_k), \quad
\mathbf{V}_{\le k} = \text{Concat}(\mathbf{V}_{\le k-1}, \mathbf{V}_k)
\end{equation}
with the cumulative cache length $T_k = \sum_{i=1}^{k} h_i w_i$ growing across scales. Attention at scale $k$ is then computed between the current-scale queries and the accumulated KV states:
\begin{equation}
\small
\mathbf{A}_k = \text{Softmax}(\mathbf{Q}_k \mathbf{K}_{\le k}^\top), \quad
\mathbf{O}_k = \mathbf{A}_k \mathbf{V}_{\le k}.
\end{equation}

As VAR generation progresses across scales, both current-scale queries and the accumulated KV states grow rapidly, leading to two major inference bottlenecks:

\noindent\textit{(i) Current-scale attention cost.} At scale $k$, the attention between $N_k$ queries and $T_k$ cached tokens incurs $\mathcal{O}(N_k T_k)$ FLOPs. Summed over all $K$ scales with $N_k = a^{2(k-1)}$, this yields an overall complexity of $\mathcal{O}(n^4)$, where $n = a^{K-1}$.

\noindent\textit{(ii) Cross-scale cache accumulation.} The cache $\{\mathbf{K}_{\le k}, \mathbf{V}_{\le k}\} \in \mathbb{R}^{T_k \times D}$ persists across all subsequent scales as context for future generation, imposing a memory footprint that grows cumulatively until the final scale $K$.

Together, these bottlenecks make efficient KV management essential for scalable deployment.

\textbf{Revisiting HACK.}
To alleviate the attention and memory inefficiency of VAR models, our
prior work 
HACK builds upon the dichotomous attention heads
in VAR models, namely \emph{contextual} heads, which mainly preserve
semantic consistency, and \emph{structural} heads, which mainly maintain
spatial coherence (detailed later in Sec.\,\ref{sec:analysis}). 
HACK first identifies
these two types of heads through an offline head classification procedure.
Throughout, we use the superscript $(p)$ with $p\in\{C,S\}$ to denote
head-type-specific variables; for instance,
$\mathbf{K}_{\le k}^{(C)},\mathbf{V}_{\le k}^{(C)}$ are the accumulated
key/value states of contextual heads, and
$\mathbf{K}_{\le k}^{(S)},\mathbf{V}_{\le k}^{(S)}$ those of structural
heads.

During inference, given an average per-head cache budget $B_c$, HACK
assigns asymmetric budgets to the two head types, denoted as
$B^{(C)}$ for contextual heads and $B^{(S)}$ for structural heads,
while keeping the average budget equal to $B_c$. For heads of type
$p\in\{C,S\}$, HACK first estimates KV's importance score
$\mathbf{S}_{\mathrm{HACK}}^{(p)}$ via a lightweight query-subset
attention mechanism, and then compresses the KV cache to the target
budget $B^{(p)}$ before attention by preserving the top-ranked KV pairs: 
\begin{equation}
\label{eq:revisit_hack}
\bar{\mathbf{K}}_{\le k}^{(p)},\,
\bar{\mathbf{V}}_{\le k}^{(p)}
=
\textsc{TopK}\!\left(
\mathbf{K}_{\le k}^{(p)},\,
\mathbf{V}_{\le k}^{(p)},\,
\mathbf{S}_{\text{HACK}}^{(p)},\,
B^{(p)}
\right),
\end{equation}
where $\bar{\mathbf{K}}_{\le k}^{(p)},\bar{\mathbf{V}}_{\le k}^{(p)}$
are the compressed KV states, which are used both for current-scale
attention computation and as the cache carried forward to subsequent
scales.

Nevertheless, HACK is limited by a coupled attention and
cache compression formulation, and by a coarse, fixed head-type budget
that lacks adaptivity across layers and generation steps. These
limitations motivate our extension, HACK++.

\begin{figure*}[t]
  \centering
  \includegraphics[width=\linewidth]{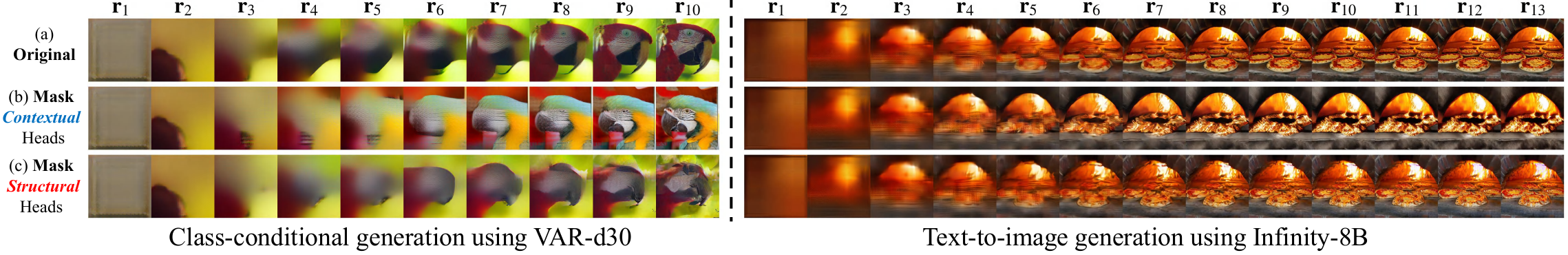}

  \caption{
\textbf{Impact of selective head masking} (15\% of total heads for each type on VAR-d30, 30\% for Infinity-8B).
Compared with the original generation (a), masking \textit{contextual} heads (b) disrupts semantic grounding, losing fine-grained content consistency with the conditioning, whereas masking \textit{structural} heads (c) retains the coarse semantic direction but causes geometric deformation and degraded spatial coherence.
}
\label{maskinghead}
\end{figure*}

\begin{figure*}[!t]
  \centering
  \includegraphics[width=\linewidth]{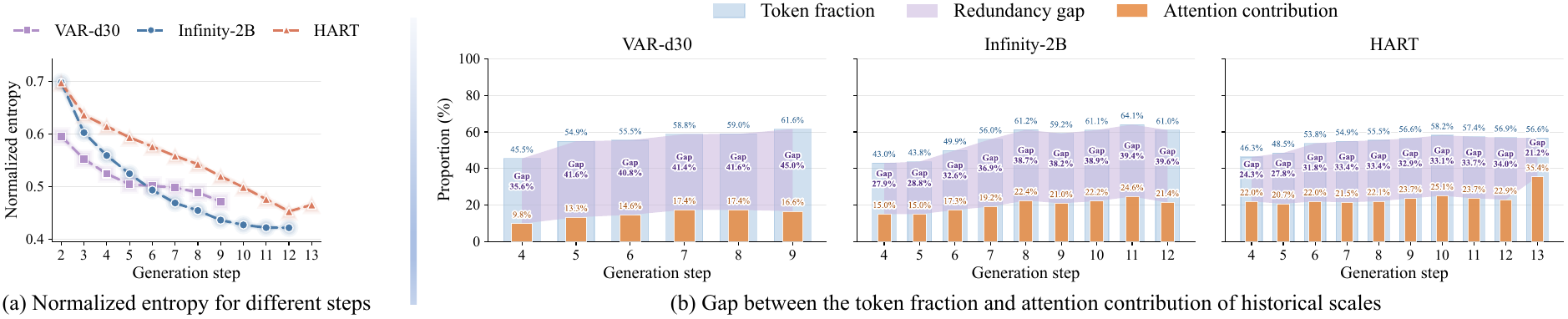}
\caption{Attention and cache require stage-decoupled compression. (a) Scale-normalized attention entropy decreases consistently across generation steps, indicating progressively more concentrated and redundant attention. (b) At later generation steps, cached historical tokens occupy a far larger fraction of the attention input than the attention mass they actually receive, exposing a substantial redundancy gap between cache storage and attention contribution.}
\label{any3}
\end{figure*}

\section{Empirical Analysis of VAR Models}
\label{sec:analysis}
Before introducing HACK++, we first conduct a systematic empirical study across
multiple VAR-family architectures, including VAR-d30, Infinity-2B/8B, and HART.
Our analysis surfaces four observations, organized into two pairs. The first
pair (Sec.\,\ref{sec:patterns},~Sec.\,\ref{sec:functional_role}) characterizes the
geometric and functional dichotomy of VAR attention heads, the foundation that
motivated our prior HACK framework. The second pair
(Sec.\,\ref{sec:compressibility},~Sec.\,\ref{sec:heterogeneity}) exposes two limitations
of HACK that the present work addresses, motivating the architectural extensions
introduced in HACK++.

\subsubsection{Dichotomous and stable attention patterns}
\label{sec:patterns}
Fig.\,\ref{obser1} visualizes attention heads across different generation scales and input samples for two representative tasks: class-conditional generation (VAR-d30) and text-to-image generation (Infinity-8B). Despite the distinct generation paradigms, they exhibit a common dichotomy: attention heads can be grouped into two geometrically distinct categories, which we term \textit{contextual heads} and \textit{structural heads}.
Contextual heads exhibit a key-centric attention pattern: most
query tokens assign high attention mass to a shared subset of
salient keys across source scales, producing vertically striped
attention maps. Importantly, these salient key positions are not
fixed across samples, but shift with the semantic content of each
input. This suggests that contextual heads primarily capture
input-dependent semantic context. In contrast, structural heads
exhibit a query-dependent spatial-correspondence pattern: each
query attends mainly to spatially adjacent tokens across preceding
scales, forming multi-diagonal attention maps. These diagonals
reflect cross-scale spatial correspondences and indicate that
structural heads maintain a stable preference over source scales.
Beyond this dichotomy, the geometric form of each pattern (vertical for
contextual heads and multi-diagonal for structural heads) remains
morphologically stable across generation scales and even across different
input samples (Fig.\,\ref{obser1}). This input-invariant consistency indicates that the contextual and structural dichotomy is an intrinsic property of pre-trained VAR models rather than an artifact of particular inputs, and suggests that head types can be reliably identified from a calibration set rather than recomputed online.

\subsubsection{Functional specialization across head types}\label{sec:functional_role}
Beyond the geometric divergence, through well-controlled head-masking experiments on both class-conditional and text-to-image generation, we verify that contextual and structural heads play
distinct functional roles during generation. 
To that effect, for each head type, we zero out the attention outputs of the same fraction of heads, while keeping others unchanged.
In Fig.\,\ref{maskinghead}, masking each
type induces a distinct dominant failure mode.
Masking contextual heads disrupts semantic grounding: the output loses
fine-grained semantic consistency, and under the richer constraints of
text-to-image generation it often fails to render the prompt-specified objects
and details. Consistent with their key-centric vertical patterns, which
anchor semantic conditioning, this motivates retaining context-relevant tokens.
Masking structural heads instead produces a spatially oriented failure mode:
the coarse semantic direction stays recognizable, but the output suffers
geometric deformation, cross-scale texture inconsistency, and degraded spatial
coherence. This aligns with their multi-diagonal patterns that
propagate geometric structure by correlating current-scale queries with
historical keys, and motivates preserving structural anchors and scale-wise spatial correspondences.
These asymmetric failure modes show that a one-size-fits-all compression
strategy is mismatched with VAR attention, motivating a pattern-specific
design whose token-selection criterion derives jointly from what each
head type must preserve and how its attention is distributed.

\begin{figure*}[!t]
  \centering
  \includegraphics[width=\linewidth]{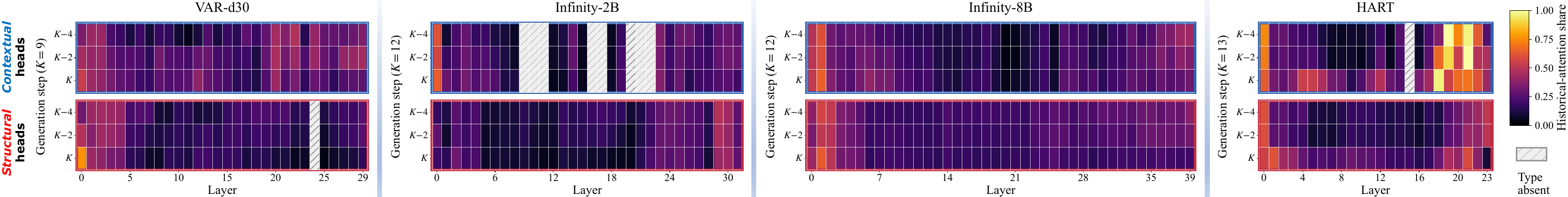}
\caption{Heterogeneous reliance on historical scales across head types, layers, and generation steps, motivating a reliance-aware three-way adaptive cache budget allocation.}
    \label{any4}
\end{figure*}

\subsubsection{Attention and cache require stage-decoupled compression}\label{sec:compressibility}
The coarse-to-fine generation paradigm makes later scales increasingly focus on
local detail refinement, so attention is more concentrated
and thus more redundant as generation proceeds. Fig.\,\ref{any3}(a) reports the
scale-normalized entropy of the attention distribution across different models,
which shows a consistent decreasing trend over generation. This aligns
with the two attention patterns in Fig.\,\ref{obser1}: contextual heads exhibit
\emph{semantic sparsity}, with attention mass dominated by a small set of
semantically salient tokens, while structural heads exhibit \emph{scale-wise
sparsity}, concentrating attention on tokens from a few preferred source scales.
Both forms of sparsity provide the basis for compact current-scale attention
computation.
The KV cache retained for future steps, however, exposes a different and even
stronger form of redundancy. At generation step $k$, the attention input
comprises tokens newly produced at the current scale and cached tokens carried
over from historical scales. To quantify how much the cache is actually used,
Fig.\,\ref{any3}(b) compares, at each step, the token fraction held by the cached
tokens with the share of attention they receive, averaged over all layers and
heads. Across all models, the cached tokens occupy a larger
proportion of the attention input than the attention mass they receive, leaving
a substantial redundancy gap that grows at later scales. Since the cache serves
all subsequent steps, this persistent over-retention indicates that it should be
compressed with a more aggressive strategy than current-step
attention.
Current-scale attention compression and future-step cache retention should
therefore not be treated as the same problem. The former serves only the
immediate generation step, whereas the latter determines what information
remains available to all subsequent steps. Applying a single shared compression
decision to both is consequently mismatched, motivating our decoupled design
(Sec.\,\ref{sec:decouple}).

\subsubsection{Heterogeneous reliance on historical scales}\label{sec:heterogeneity}
Although redundant historical tokens admit aggressive
compression, this redundancy is far from uniform: different head
types, across layers and generation steps, rely on historical tokens to highly
heterogeneous degrees. To characterize this, we compute for each head its
\emph{historical-attention share}, the fraction of its attention directed to
historical tokens from preceding scales out of all attention it assigns to
historical and current-scale tokens. We aggregate this share by head type within
each layer, and Fig.\,\ref{any4} visualizes the resulting
(layer, head-type) shares across later generation steps, revealing heterogeneity
along three dimensions. \textit{First}, within a layer, contextual and structural heads
can rely on history to very different degrees, so the budget should adapt to
head type.  \textit{Second}, within a head type, reliance varies markedly across layers,
so the budget should adapt to depth. \textit{Third}, for a given head, reliance evolves
as generation proceeds, so the budget assigned at step $k$ must anticipate
future-step needs. Because reliance is heterogeneous along head type, layer, and
generation step simultaneously, a single static budget is inadequate; this
motivates the reliance-aware, three-way adaptive budget allocation of HACK++
(Sec.\,\ref{sec:budget}).

\begin{figure*}[t]
  \centering
  
  \includegraphics[width=1.0\linewidth]{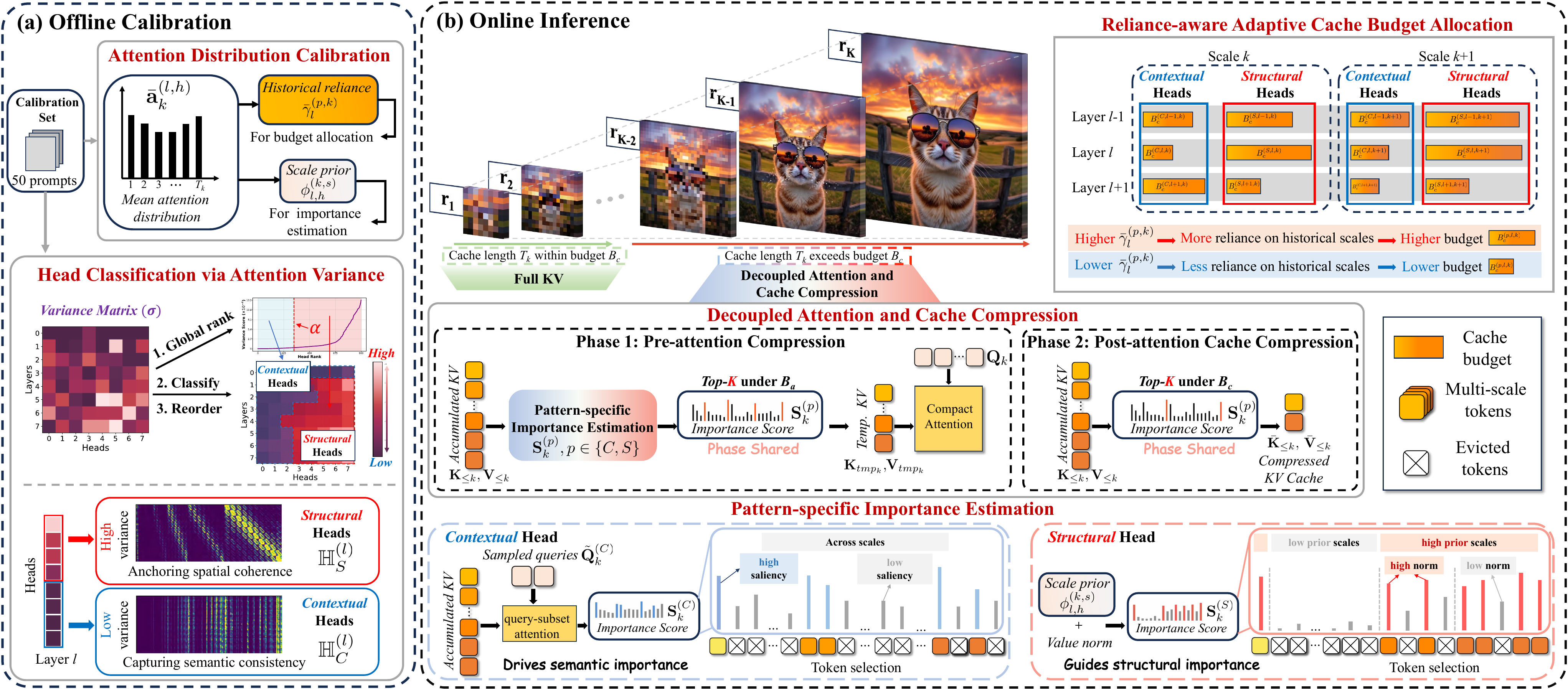}
\caption{\textbf{Overview of the HACK++ framework.}
\textbf{(a) Offline Calibration:} On a small calibration set,
HACK++ computes per-head, per-scale mean attention distributions to derive scale priors and historical reliance scores (Sec.\,\ref{sec:attn_calib}) and classifies all heads into \textit{contextual heads} and \textit{structural heads} via attention variance ranking (Sec.\,\ref{sec:head_cls}).
\textbf{(b) Online Inference:} At each scale $k$ where the 
accumulated KV length exceeds the budget, HACK++ applies 
decoupled two-phase compression (Sec.\,\ref{sec:decouple}): pre-attention compression 
bounds attention cost within budget $B_a$, and post-attention 
cache compression independently retains the most important 
tokens within budget $B_c$. Both phases are driven by 
pattern-specific importance estimation tailored to each head 
type (Sec.\,\ref{sec:scoring}), with cache budgets adaptively allocated via historical reliance (Sec.\,\ref{sec:budget}).
}
\label{overall}
\end{figure*}

\section{Methodology}
\label{sec:method}
This section formally presents HACK++ (overall framework in
Fig.\,\ref{overall}).
Guided by the empirical findings in Sec.\,\ref{sec:analysis}, it
operates in two stages.
\textbf{1. Offline Calibration.}
From a small calibration set, HACK++ extracts two lightweight
primitives: per-head, per-scale mean attention distributions
$\bar{\mathbf{a}}_k^{(l,h)}$ (Sec.\,\ref{sec:attn_calib}) and a binary
contextual/structural head classification via attention variance (Sec.\,\ref{sec:head_cls}). 
\textbf{2. Online Inference.}
At each scale $k$, HACK++ applies a \emph{decoupled two-phase
compression} (Sec.\,\ref{sec:decouple}): a pre-attention phase bounds the
current-scale attention cost under budget $B_a$, and a
post-attention phase compresses the accumulated states under an
adaptively allocated cache budget (Sec.\,\ref{sec:budget}). Both phases select tokens via the same \emph{pattern-specific importance estimation} (Sec.\,\ref{sec:scoring}) computed per head type.

\subsection{Offline Calibration}
\label{sec:calibration}

As shown in Sec.\,\ref{sec:patterns}, attention patterns in
VAR models remain stable across input samples and
generation scales. HACK++ exploits this stability through a one-time offline calibration
pipeline (Fig.\,\ref{overall}(a)), which extracts
the metadata required for online compression from a
small calibration set.

\subsubsection{Attention Distribution Calibration}
\label{sec:attn_calib}

We first collect the mean attention distribution across
scales to capture each head's per-scale attention
preference. During the calibration forward passes, for
each head $h$ in layer $l$ at generation step $k$, we
record the attention matrix
$\mathbf{A}_k^{(l,h)} \in \mathbb{R}^{N_k \times T_k}$
and compute the mean attention distribution by averaging
over queries and calibration samples:
\begin{equation}
\label{eq:mean_attn}
\bar{\mathbf{a}}_{k}^{(l,h)}
= \mathbb{E}_{\text{cal}}
\left[\,
\frac{1}{N_k}
\sum_{i=1}^{N_k}
\mathbf{A}_{k}^{(l,h)}[i,\,:]
\,\right]
\in \mathbb{R}^{T_k},
\end{equation}
where $\mathbb{E}_{\text{cal}}[\cdot]$ denotes averaging
over the calibration set.
Intuitively, $\bar{\mathbf{a}}_{k}^{(l,h)}[j]$ measures the
average attention that head $(l,h)$ allocates to the
$j$-th KV token when generating at scale $k$, revealing
which scales and which tokens each head preferentially
attends to.

The mean attention distribution is the common
foundation for two derived quantities introduced later
in the pipeline: the \emph{scale-prior factor}
$\phi_{l,h}^{(k,s)}$ that serve as a cross-scale prior
in structural head scoring
(Sec.\,\ref{sec:scoring_structural}), and the
\emph{history reliance ratios} $\gamma_{l,h}^{(k)}$
that drive adaptive cache budget allocation
(Sec.\,\ref{sec:budget}). Both are defined at their points
of use to keep the calibration stage minimal.

\subsubsection{Head Classification via Attention Variance}
\label{sec:head_cls}

We classify each head as contextual or structural head based
on its attention variance. As in
Sec.\,\ref{sec:functional_role}, contextual heads
selectively focus on a small set of semantically salient
tokens regardless of query position, yielding low
column-wise variance; structural heads attend in a position-sensitive manner that shifts with
spatial configurations, resulting in higher variance.

To quantify this divergence, we analyze the attention
matrix $\mathbf{A}_K^{(l,h)}$ at the final generation
scale $K$, which encapsulates interactions from all
prior scales. For each head $h$ in layer $l$, we compute
the sum of column-wise variances and average over the
calibration set:
\begin{equation}
\label{eq:variance}
\sigma_{l,h} = \mathbb{E}_{\text{cal}}
\left[\,
\sum_{j=1}^{T_K}
\mathrm{Var}\!\left(
\mathbf{A}_{K}^{(l,h)}[:,\, j]
\right)
\,\right].
\end{equation}

This yields a variance matrix
${\sigma} \in \mathbb{R}^{L \times H}$, where $L$
and $H$ denote the number of layers and heads per
layer, respectively.
We globally rank all $L \times H$ heads by
$\sigma_{l,h}$: the bottom-$\alpha$ fraction (low
variance, vertically concentrated patterns) are
classified as \emph{contextual heads}, and the
remainder (high variance, multi-diagonal patterns) as
\emph{structural heads}, where $\alpha$ is the global
contextual head ratio tuned to each model's empirical
variance distribution.

Following classification, we obtain per-layer index sets
$\mathbb{H}_C^{(l)}$ and $\mathbb{H}_S^{(l)}$ for
contextual and structural heads in layer $l$,
satisfying
$\mathbb{H}_C^{(l)} \cup \mathbb{H}_S^{(l)}
= \{1, \dots, H\}$
and
$\mathbb{H}_C^{(l)} \cap \mathbb{H}_S^{(l)}
= \emptyset$.
To facilitate efficient inference, we reorder heads within each layer by grouping them according to type.

\subsection{Decoupled Attention and Cache Compression} \label{sec:decouple}
As analyzed in Sec.\,\ref{sec:compressibility}, the attention computation
serves only the current scale, whereas the KV cache is stored for all
subsequent scales; this stage mismatch leads to their divergent
compressibility. HACK conflates the two objectives into a single pre-attention operation, imposing a rigid coupling that forces one budget to serve both purposes and thereby preventing aggressive cache compression. Attention and cache should instead be decoupled, so that each is compressed in accordance with the scale it actually serves.

HACK++ resolves this tension with two separate compression steps. The \emph{pre-attention compression} explicitly bounds the growth of attention computation, retaining a compact set of tokens that reduces the current-scale attention cost while preserving enough context for accurate generation. The \emph{post-attention cache compression} compresses the accumulated KV states stored for future scales, forming a minimal cache that governs both the long-term memory footprint and the historical information available to subsequent steps. 

\textbf{Phase~1: Pre-attention compression.}
When the accumulated length $T_k$ exceeds the attention budget $B_a$,
HACK++ compresses the KV states before attention. For each head
$h \in \mathbb{H}_p^{(l)}$, $p \in \{C, S\}$, it computes the
importance score $\mathbf{S}_k^{(p)} \in \mathbb{R}^{T_k}$ via the
pattern-specific estimation strategy $f_p$ (Sec.\,\ref{sec:scoring}), and
selects the top-$B_a$ tokens from the accumulated states
$\mathbf{K}_{\le k}^{(p)}, \mathbf{V}_{\le k}^{(p)}$ (the cache retained
from the previous step concatenated with the current scale's tokens)
to form a \emph{temporary} compact subset solely for current-scale
attention:
\begin{equation}
\label{eq:attn_compress}
\mathbf{K}_{tmp_k}^{(p)},\,
\mathbf{V}_{tmp_k}^{(p)}
= \textsc{TopK}\!\left(
\mathbf{K}_{\le k}^{(p)},\,
\mathbf{V}_{\le k}^{(p)},\,
\mathbf{S}_k^{(p)},\, B_a
\right).
\end{equation}

Attention is then computed using the temporary KV states: 
\begin{equation}
\label{eq:attn_split}
\mathbf{O}_k^{(p)} = \text{Softmax}\!\left(
\mathbf{Q}_k^{(p)}\,
(\mathbf{K}_{tmp_k}^{(p)})^\top
\right)
\mathbf{V}_{tmp_k}^{(p)}.
\end{equation}

This subset is used only for the current step and then discarded, so
$B_a$ acts purely as a compute throttle, bounding the per-step
attention cost without affecting what is stored.

\textbf{Phase~2: Post-attention cache compression.} After attention is computed, HACK++ further selects the KV pairs to be cached for subsequent scales. Given an average cache budget $B_c$, HACK++ selects, for each head type, the top-$B_c^{(p,l,k)}$ tokens from the accumulated KV states $\mathbf{K}_{\le k}^{(p)}, \mathbf{V}_{\le k}^{(p)}$ to form the stored cache, reusing the Phase~1 scores $\mathbf{S}_k^{(p)}$ at no additional cost: \begin{equation}\label{eq:cache_compress} \bar{\mathbf{K}}_{\le k}^{(p)},\, \bar{\mathbf{V}}_{\le k}^{(p)} = \textsc{TopK}\!\left(\mathbf{K}_{\le k}^{(p)},\, \mathbf{V}_{\le k}^{(p)},\, \mathbf{S}_k^{(p)},\, B_c^{(p,l,k)} \right), \end{equation}where the per-(layer, type, step) budget $B_c^{(p,l,k)}$ is the adaptive allocation derived from the average target $B_c$ (Sec.\,\ref{sec:budget}).

 In this way, attention computation and cache
compression are decoupled, so that their token selections are
governed by independent budgets and do not constrain each other.

\subsection{Pattern-Specific Importance Estimation}
\label{sec:scoring}
The two head types exhibit different attention patterns
(Sec.\,\ref{sec:patterns}) and functional roles
(Sec.\,\ref{sec:functional_role}), so no single importance criterion can
faithfully identify the critical tokens for both. To address this, we
develop a pattern-specific importance estimation strategy that adapts
token selection to the distinct behaviors of contextual and structural
heads.

\subsubsection{Contextual-Head Importance Estimation}
\label{sec:scoring_contextual}

Contextual heads exhibit vertically concentrated
attention patterns, selectively attending to a small set
of semantically salient tokens across scales. Since
these salient tokens are content-dependent and vary
across inputs, contextual heads require online
attention-guided selection. However, relying on full attention for token selection
contradicts the goal of attention compression, motivating
a lightweight online approximation.

To this end, we employ a \emph{query-subset attention}
strategy. The key observation is that contextual heads
exhibit query-consistent attention: different queries
tend to attend to a shared set of salient semantic
tokens, so a small subset of queries suffices to
identify the important KV pairs. For each contextual
head, we uniformly sample $N_{\text{obs}}$ queries from
the current scale's $N_k$ queries, obtaining
$\tilde{\mathbf{Q}}_k^{(C)}
\in \mathbb{R}^{N_{\text{obs}} \times D}$, and
compute approximate attention scores over the
accumulated contextual KV cache:
\begin{equation}
\label{eq:subset_attn}
\tilde{\mathbf{A}}_k^{(C)} =
\text{Softmax}\!\left(
\tilde{\mathbf{Q}}_k^{(C)}
(\mathbf{K}_{\le k}^{(C)})^\top
\right).
\end{equation}

The importance score for each KV pair is then
estimated by averaging the attention scores across the
sampled queries, followed by a local max-pooling
operation:
\begin{equation}
\label{eq:ctx_score}
\mathbf{S}_k^{(C)}[j] =
\text{MaxPool}\!\left(
\frac{1}{N_{\text{obs}}}
\sum_{i=1}^{N_{\text{obs}}}
\tilde{\textbf{A}}_{k}^{(C)}[i, j]
\right).
\end{equation}

The max-pooling operation promotes spatially coherent
token selection by propagating high scores to
neighboring positions, ensuring that semantically
salient regions are retained as contiguous groups
rather than isolated tokens~\cite{li2024snapkv}.
Positions with higher $\mathbf{S}_k^{(C)}[j]$ are
prioritized for retention during compression.
We refer to this contextual-head importance estimation 
strategy collectively as $f_C$.
As shown in Fig.\,\ref{distribution}(a,b),
query-subset attention largely preserves the salient
attention peaks of contextual heads, enabling efficient
and accurate importance estimation.

\subsubsection{Structural-Head Importance Estimation}
\label{sec:scoring_structural}
\begin{figure}[t]
  \centering
      \includegraphics[width=0.5\textwidth]{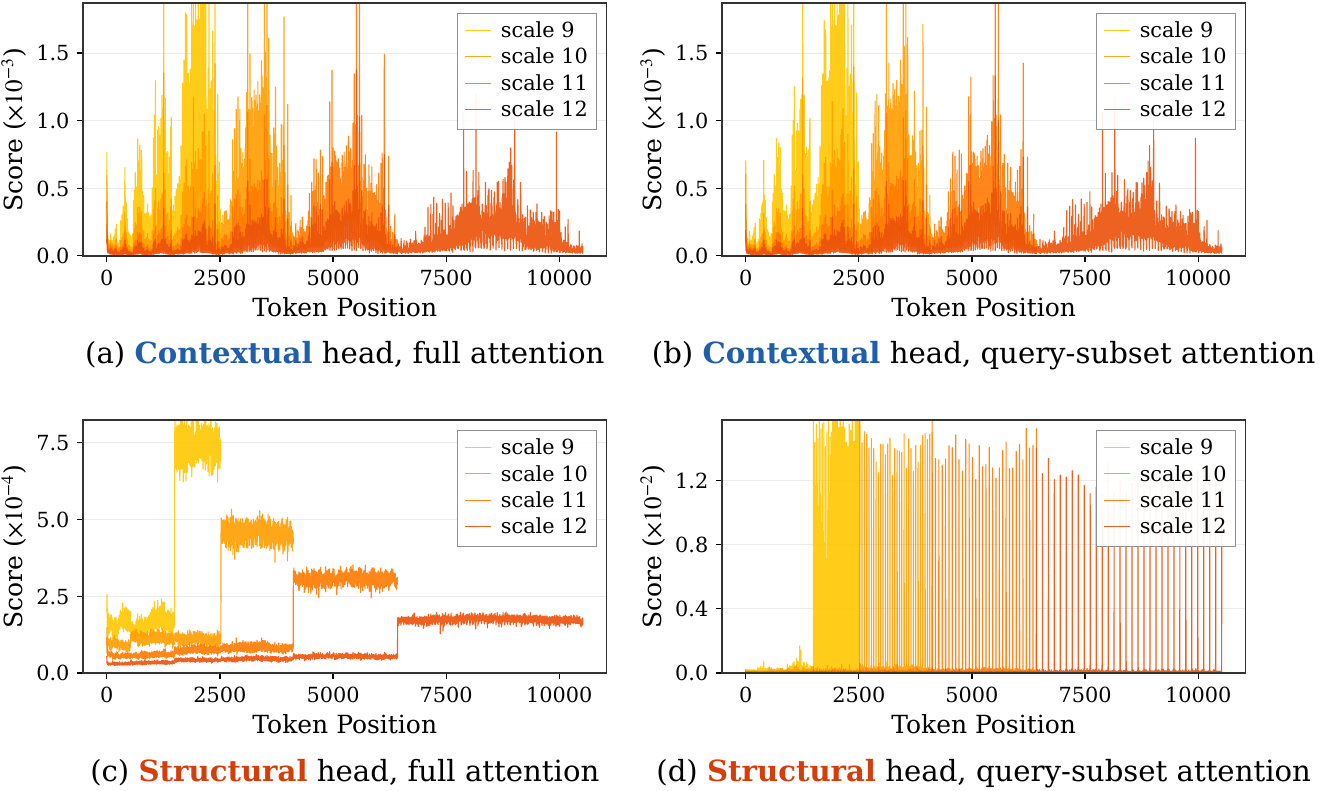}
\caption{Comparison of mean attention distributions under full
and query-subset attention on the Infinity model.
(a, b)~For contextual heads, query-subset attention faithfully
preserves the salient peaks of full attention.
(c, d)~For structural heads, it instead degenerates into a
near-uniform within-scale distribution, failing to identify
informative tokens within each scale.
Different colors denote different generation scales.}
\label{distribution}
\end{figure}

Unlike contextual heads, structural heads exhibit scale-preferential,
query-dependent attention, where each query primarily attends to tokens
at spatially corresponding positions across scales. Query-subset
attention is therefore ill-suited here: as illustrated in
Fig.\,\ref{distribution}(c,d), although it can roughly localize the
preferred scale range, each sampled query highlights only its own
spatially corresponding tokens, so the averaged scores merely reflect
the sampled query positions and token selection degenerates into
near-uniform sampling within each scale, failing to preserve the
structurally informative anchors.

To address this issue, we decompose structural-head importance
estimation into a \emph{cross-scale prior} that weights source scales
by their structural relevance, and a \emph{within-scale salience} term
that selects informative anchors within those scales.
Building on the pattern stability established in
Sec.\,\ref{sec:patterns}, we aggregate the calibrated mean attention
distribution $\bar{\mathbf{a}}_{k}^{(l,h)}$ from
Sec.\,\ref{sec:attn_calib} to the scale level. Specifically, at each scale
$k$, for each structural head $h$ in layer $l$, we define its
scale-prior factor for each source scale $s \in \{1,\ldots,k\}$
as:
\begin{equation}
\label{eq:scale_factor}
\phi_{l,h}^{(k,s)} =
\frac{1}{N_s}
\sum_{j \in \mathcal{T}_s}
\bar{\mathbf{a}}_{k}^{(l,h)}[j],
\end{equation}
where $\mathcal{T}_s$ denotes the set of token indices belonging to
scale $s$ and $N_s = |\mathcal{T}_s|$. A larger $\phi_{l,h}^{(k,s)}$
indicates that head $(l,h)$ relies more strongly on source scale $s$
when generating at scale $k$. These factors are computed once during
calibration and stored as head-level metadata with negligible
overhead. We then assign each token the calibrated factor of its
source scale via the weighting function
$\omega_{l,h}^{(k)}: \{1,\ldots,T_k\} \to \mathbb{R}^{+}$,
\begin{equation}
\label{eq:scale_weight}
\omega_{l,h}^{(k)}(j) =
\phi_{l,h}^{(k, s(j))},
\end{equation}
where $s(j)$ denotes the source scale to which token $j$ belongs, and
compute the structural KV pairs' importance score as:
\begin{equation}
\label{eq:struct_score}
\mathbf{S}_k^{(S)}[j] =
\omega_{l,h}^{(k)}(j) \cdot
\left\| \mathbf{V}_{\le k}^{(S)}[j] \right\|_2 .
\end{equation}
The two terms are complementary. The calibrated weight
$\omega_{l,h}^{(k)}$ acts as a cross-scale prior, steering selection
toward the source scales each head relies on; the value norm $\|\mathbf{V}_{\le k}^{(S)}[j]\|_2$ measures within-scale salience, as tokens with larger norm contribute more to the
attention output. Together they select high-impact structural anchors from
the most relevant scales. We refer to this strategy collectively as
$f_S$. Notably, $f_S$ requires no query--key attention: the
cross-scale prior $\phi$ is precomputed offline, leaving only a cheap
value norm to be evaluated online, so structural-head scoring is
essentially free.

\subsection{Reliance-Aware Adaptive Cache Budget Allocation}
\label{sec:budget}

To address the heterogeneous reliance on historical
scales, we develop an adaptive cache budget allocation strategy that distributes the budget across head types,
layers, and generation steps according to each group's
actual demand on historical tokens. A central observation
is that the cache retained after step $k$ is consumed
not by step $k$ itself, but by all subsequent steps
$\{k{+}1,\ldots,K\}$, which repeatedly attend back to the history tokens it stores. The budget assigned to a cache group should therefore reflect how strongly these future steps will rely on historical scales.
We quantify the historical reliance of each head by
measuring the average contribution of its cache across
subsequent generation steps. For head $(l,h)$ at step
$k$, this is computed as:
\begin{equation}
\label{eq:hist_reliance_head}
\gamma_{l,h}^{(k)}
=
\sum_{k'=k+1}^{K}
\frac{1}{|\mathcal{T}_{<k'}|}
\sum_{j\in\mathcal{T}_{<k'}}
\bar{\mathbf{a}}_{k'}^{(l,h)}[j],
\end{equation}
where $\mathcal{T}_{<k'}$ denotes the tokens from scales preceding
$k'$. The inner term is the average attention each historical token
receives at step $k'$, and summing over $k' > k$
accumulates these contributions across all future
consumers of the cache. Aggregating within each (layer,
head-type) group gives:
\begin{equation}
\label{eq:hist_reliance_group}
\bar{\gamma}_{l}^{(p,k)}
=
\frac{1}{|\mathbb{H}_{p}^{(l)}|}
\sum_{h\in\mathbb{H}_{p}^{(l)}}
\gamma_{l,h}^{(k)},
\qquad p\in\{C,S\},
\end{equation}
where a larger $\bar{\gamma}_{l}^{(p,k)}$ indicates
that heads of type $p$ in layer $l$ rely more strongly
on historical cache during the remaining generation.
HACK++ then distributes the cache budget according to
this historical reliance score:
\begin{equation}
\label{eq:cache_alloc}
\begin{gathered}
B_c^{(p,l,k)}
\propto
\left(\bar{\gamma}_{l}^{(p,k)}\right)^{\tau},
\\[3pt]
\text{s.t.}\quad
\frac{1}{LH}
\sum_{l=1}^{L}\left(
\sum_{p\in\{C,S\}}
|\mathbb{H}_{p}^{(l)}| \cdot B_c^{(p,l,k)}\right)
=
B_c .
\end{gathered}
\end{equation}
where $B_c$ is the target average per-head cache budget, and coefficient $\tau$ controls the sharpness of the allocation. 

In this way, HACK++ achieves adaptive budget allocation
across head types, layers, and generation steps, naturally
capturing the three-axis heterogeneity.
\subsection{Complexity Analysis}

\noindent\textbf{Attention complexity.}
Vanilla VAR attends over all $T_k$ cached tokens with $N_k$ queries at each scale $k$, incurring a total attention complexity of $\mathcal{O}(n^4)$ due to cumulative KV cache growth. Like HACK, HACK++ bounds the attention KV length within the attention budget $B_a$, reducing the per-step cost to $N_k \cdot \min(T_k, B_a)$, where $N_k$ is the number of tokens at step $k$ and $T_k$ is the cumulative cache length up to step $k$. Summing over all scales yields:
\begin{equation}
\small
\sum_{k=1}^{K} N_k \cdot \min(T_k, B_a)
\;\leq\; B_a \sum_{k=1}^{K} a^{2(k-1)}
\;\sim\; \mathcal{O}(B_a n^2),
\end{equation}
matching the asymptotic reduction of HACK while explicitly decoupling the attention budget from the cache budget.

\noindent\textbf{KV cache capacity.}
The cache budget $B_c$ independently controls the stored cache, bounding it to $B_c$ per head once compression triggers and yielding a total cached length of $B_c \cdot H \cdot L$. Whereas HACK couples the two budgets ($B_c = B_a$), HACK++ decouples them and sets $B_c < B_a$, enabling far more aggressive cache compression without inflating the attention budget.

\noindent\textbf{Compression overhead.}
The overhead in HACK arises from subset attention applied to all heads, where $N_{\text{obs}} \ll N_k$ sampled queries estimate token importance. HACK++ incurs lower overhead through its enhanced, attention-independent structural-head scoring: the precomputed scale factor combined with a single-pass value-norm reduction eliminates the $N_{\text{obs}}$ factor on the $(1{-}\alpha)H$ structural heads. The compression overhead of HACK++ decomposes by head type as:
\begin{equation}
\underbrace{\alpha H \cdot N_{\text{obs}} \cdot T_k}_{\text{contextual (subset attn)}}
+
\underbrace{(1 - \alpha) H \cdot T_k}_{\text{structural (v-norm only)}}
\;<\;
\underbrace{H \cdot N_{\text{obs}} \cdot T_k}_{\text{HACK (subset attn, all heads)}},
\end{equation}
where the right-hand side recovers HACK's overhead.


 

\begin{table*}[t]
\centering
\caption{Quantitative comparison of text-to-image generation on
Infinity-2B/8B~\cite{han2024infinity} and HART~\cite{tang2024hart}.
Per-category and overall FID/CLIP are computed on
MJHQ-30K~\cite{li2024playground}. HPSv2.1~\cite{wu2023human}
subcategories are abbreviated as Painting (Pai.), Anime (Ani.),
Photo (Pho.), and Concept Art (Art). KV size and attention TFLOPs are
measured at batch size 1.}
\label{tab:t2i}
\setlength{\tabcolsep}{4pt}
\resizebox{\textwidth}{!}{%
\begin{tabular}{l cc | cc | cc cc cc cc cc | ccccc | c}
\toprule
\multirow{3}{*}{Method}
 & \multirow{3}{*}{$\eta_\text{a}$}
 & \multirow{3}{*}{$\eta_\text{c}$}
 & \multirow{3}{*}{\makecell{KV \\Size$\downarrow$}}
 & \multirow{3}{*}{\makecell{Attn.\\TFLOPs$\downarrow$}}
 & \multicolumn{2}{c}{Plants}
 & \multicolumn{2}{c}{Food}
 & \multicolumn{2}{c}{Landscape}
 & \multicolumn{2}{c}{People}
 & \multicolumn{2}{c|}{Overall}
 & \multicolumn{5}{c|}{HPSv2.1$\uparrow$}
 & \multirow{3}{*}{IR$\uparrow$} \\
\cmidrule(lr){6-7}\cmidrule(lr){8-9}\cmidrule(lr){10-11}\cmidrule(lr){12-13}\cmidrule(lr){14-15}\cmidrule(lr){16-20}
 &&&&& \scriptsize{FID$\downarrow$} & \scriptsize{CLIP$\uparrow$}
     & \scriptsize{FID$\downarrow$} & \scriptsize{CLIP$\uparrow$}
     & \scriptsize{FID$\downarrow$} & \scriptsize{CLIP$\uparrow$}
     & \scriptsize{FID$\downarrow$} & \scriptsize{CLIP$\uparrow$}
     & \scriptsize{FID$\downarrow$} & \scriptsize{CLIP$\uparrow$}
     & Pai. & Ani. & Pho. & Art. & Avg. & \\
\midrule
\multicolumn{21}{l}{\textit{Infinity-2B}~\cite{han2024infinity}} \\
\midrule
Full         & 100\% & 100\% & 7.71\,GB & 35.91
 & 30.96 & 26.33 & 31.68 & 26.64 & 25.98 & 26.15 & 31.62 & 30.26 & 10.34 & 27.52 & 30.48 & 31.64 & 29.41 & 30.43 & 30.49 & 0.946 \\
Streaming~\cite{xiao2023efficient}   & 30\%  & 30\%  & 2.31\,GB & 15.34 & 32.59 & 26.16 & 33.67 & 26.63 & 27.04 & 26.17 & 31.89 & 28.17 & 11.20 & 27.54 & 29.58 & 30.92 & 28.96 & 29.60 & 29.76 & 0.901 \\
H2O~\cite{zhang2024h2o}         & 30\%  & 30\%  & 2.31\,GB & 15.34 & 31.72 & 26.29 & 32.59 & 26.71 & 26.62 & 26.26 & 30.99 & 28.13 & 10.68 & 27.57 & 29.46 & 30.79 & 28.69 & 29.48 & 29.60 & 0.910 \\
SnapKV~\cite{li2024snapkv}      & 30\%  & 30\%  & 2.31\,GB & 15.34 & 30.83 & 26.30 & 32.47 & 26.79 & 25.46 & 26.30 & 30.62 & 28.13 & 10.60 & 27.56 & 29.48 & 30.80 & 28.61 & 29.53 & 29.60 & 0.904 \\
LOOK-M~\cite{wan2024look}      & 30\%  & 30\%  & 2.31\,GB & 15.34 & 32.53 & 26.26 & 33.20 & 26.91 & 26.49 & 26.34 & 31.36 & 28.21 & 11.14 & 27.59 & 28.51 & 29.90 & 27.81 & 28.70 & 28.73 & 0.864 \\
CAKE~\cite{qin2025cake}        & 30\%  & 30\%  & 2.31\,GB & 15.34 & 31.67 & 26.25 & 32.49 & 26.77 & 25.94 & 26.30 & 31.39 & 28.16 & 10.59 & 27.56 & 29.30 & 30.64 & 28.47 & 29.41 & 29.46 & 0.906 \\
MEDA~\cite{wan2025meda}        & 30\%  & 30\%  & 2.31\,GB & 15.34 & 32.47 & 26.26 & 33.02 & 26.91 & 26.41 & 26.35 & 31.44 & 28.20 & 11.14 & 27.59 & 28.50 & 29.92 & 27.76 & 28.62 & 28.70 & 0.867 \\
HACK~\cite{qin2026head}        & 30\%  & 30\%  & 2.31\,GB & 15.34 & 31.54 & 26.24 & 31.95 & 26.79 & 26.14 & 26.29 & 31.61 & 28.23 & 10.56 & 27.62 & 30.10 & 31.35 & 29.13 & 30.13 & 30.18 & 0.933 \\

ScaleKV~\cite{li2025scalekv}      & --  & 10\%  & 0.77\,GB & 19.19 & 31.53 & 26.28 & 32.30 & 26.67 & 25.95 & 26.13 & 31.53 & 28.04 & 10.55 & 27.54 & 30.32    & 31.50    & 29.35    & 30.29    & 30.37    & 0.942    \\
ScaleKV~\cite{li2025scalekv}      & --  & 1\%  & 0.08\,GB & 14.35 &32.87  & 26.14 & 33.36 & 26.51 & 27.97 & 26.09 & 31.41 & 28.01 & 10.99 & 27.47 & 29.83   & 31.27    & 29.23    & 29.88    & 30.05   & 0.908   \\
\ours HACK++ & 30\% & 10\%  & 0.77\,GB & 14.32
 & 30.78 & 26.27 & 31.95 & 26.69 & 26.01 & 26.11 & 31.27 & 28.06 & 10.49 & 27.54 & 30.44    & 31.70    & 29.45    & 30.41    & 30.50    & 0.953    \\
\ours HACK++ & 30\% & 1\%   & 0.08\,GB & 12.11 & 31.39 & 26.16 & 32.17 & 26.66 & 26.67 & 26.10 & 31.04 & 28.03 & 10.50 & 27.50 &  30.09 & 31.51 & 29.27   & 30.13   & 30.25    & 0.934    \\
\midrule
\multicolumn{21}{l}{\textit{Infinity-8B}~\cite{han2024infinity}} \\
\midrule
Full         & 100\% & 100\% & 16.86\,GB & 78.54
 & 28.81 & 28.12 & 28.61 & 28.45 & 24.58 & 27.32 & 30.26 & 29.42 & 8.75 & 29.10 & 30.73 & 32.44 & 29.48 & 31.31 & 30.99 & 1.049 \\
Streaming~\cite{xiao2023efficient}   & 30\%  & 30\%    & 5.06\,GB & 33.55 & 29.70 & 28.02 & 28.48 & 28.47 & 23.88 & 27.34 & 30.49 & 29.40 & 8.98 &29.05 & 30.29 & 31.90 & 29.14 & 30.75 & 30.52 & 1.016 \\
H2O~\cite{zhang2024h2o}         & 30\%  & 30\%   & 5.06\,GB & 33.55 & 29.55 & 28.01 & 29.19 & 28.41 & 23.78 & 27.45 & 30.43 & 29.36 & 9.04 & 29.02 & 30.31 & 32.01 & 29.04 & 30.77 & 30.53 & 1.020 \\
SnapKV~\cite{li2024snapkv}      & 30\%  & 30\%   & 5.06\,GB & 33.55 & 30.23 & 28.01 & 29.64 & 28.45 & 23.23 & 27.58 & 30.65 & 29.46 & 9.45 & 29.08 & 29.96 & 31.72 & 28.70 & 30.44 & 30.21 & 1.015 \\
LOOK-M~\cite{wan2024look}      & 30\%  & 30\%   & 5.06\,GB & 33.55 & 32.29 & 27.89 & 30.79 & 28.50 & 24.55 & 27.55 & 32.81 & 29.45 & 10.84 & 29.04 & 29.70 & 31.53 & 28.60 & 30.13 & 29.99 & 0.994 \\
CAKE~\cite{qin2025cake}         & 30\%  & 30\%   & 5.06\,GB & 33.55 & 30.74 & 27.86 & 30.06 & 28.44 & 23.38 & 27.54 & 31.19 & 29.44 & 9.80 & 29.05 & 29.79 & 31.55 & 28.59 & 30.25 & 30.04 & 1.002 \\
MEDA~\cite{wan2025meda}        & 30\%  & 30\%   & 5.06\,GB & 33.55 & 33.48 & 28.01 & 32.14 & 28.42 & 25.11 & 27.43 & 33.20 & 29.41 & 11.56 & 29.00 & 29.30 & 31.13 & 28.14 & 29.71 & 29.57 & 0.954 \\
HACK~\cite{qin2026head}         & 30\%  & 30\%  & 5.06\,GB & 33.54 & 28.72 & 28.01 & 28.71 & 28.42 & 23.28 & 27.43 & 29.94 & 29.42 & 8.62 & 29.08 & 30.53 & 32.16 & 29.11 & 30.96 & 30.69 & 1.043 \\

ScaleKV~\cite{li2025scalekv}   & --  & 10\%  & 1.69\,GB & 41.98 & 28.97 & 28.10 & 29.12 & 28.39 & 24.51 & 27.30 & 30.01 & 29.35 & 8.78 & 29.06 & 30.66    & 32.34    & 29.33    & 31.18    & 30.88    & 1.042    \\

ScaleKV~\cite{li2025scalekv}      & --  & 1\%  & 0.17\,GB & 31.40 & 30.57 & 28.08 & 28.49 & 28.74 & 23.92 & 27.42 & 31.33 & 29.61 & 9.49 & 29.13 & 29.43    & 31.03    & 28.69    & 29.92   & 29.77    & 0.972    \\

\ours HACK++ & 30\% & 10\%  & 1.69\,GB & 31.53 & 28.82 & 28.09 & 28.57 & 28.47 & 24.00 & 27.39 & 30.10 & 29.44 & 8.56 & 29.11 & 30.73    & 32.39    & 29.31    & 31.20   &  30.91  & 1.054   \\
\ours HACK++ & 30\% & 1\%   & 0.17\,GB & 26.50 & 30.09 & 28.08 &  29.66 & 28.47 & 24.39 & 27.42 & 30.72 & 29.41 & 9.11 & 29.09    & 30.05    & 31.70    & 29.01    & 30.62   & 30.35    & 1.015    \\
\midrule
\multicolumn{21}{l}{\textit{HART}\cite{tang2024hart}} \\
\midrule
Full         & 100\% & 100\% & 2.68\,GB & 17.66 & 30.47 & 27.25 & 30.67 & 27.26 & 24.53 &26.14  & 30.15 & 27.90 & 10.70 & 27.69 & 28.58  & 30.57   & 27.53   & 28.67    & 28.75 & 0.658 \\
Streaming~\cite{xiao2023efficient}   & 30\%  & 30\% & 0.80\,GB & 7.40 & 32.13 & 26.96 & 31.84 & 27.41 & 23.22 & 26.12 & 30.37 & 27.43 & 11.16 & 27.38 & 25.16 & 26.98 & 24.63 & 25.50 & 25.57 & 0.458 \\
H2O~\cite{zhang2024h2o}         & 30\%  & 30\%  & 0.80\,GB & 7.40 & 31.49 & 26.51 & 33.78 & 27.28 & 22.42 & 26.07 & 31.97 & 27.64 & 11.13 & 27.32 & 25.52 & 27.42 & 24.25 & 25.58 & 25.69 & 0.475 \\
SnapKV~\cite{li2024snapkv}       & 30\%  & 30\%  & 0.80\,GB & 7.40 & 31.71 & 26.51 & 33.87 & 27.34 & 22.44 & 26.16 & 32.36 & 27.62 & 11.33 & 27.33 & 25.49 & 27.17 & 24.26 & 25.54 & 25.62 & 0.469 \\
LOOK-M~\cite{wan2024look}      & 30\%  & 30\%  & 0.80\,GB & 7.40 & 44.69 & 24.87 & 48.21 & 26.04 & 41.37 & 24.77 & 51.78 & 26.13 & 25.64 & 25.85 & 20.99 & 21.33 & 19.87 & 21.15 & 20.83 & 0.140 \\
CAKE~\cite{qin2025cake}        & 30\%  & 30\%  & 0.80\,GB & 7.40 & 32.75 & 26.40 & 36.27 & 27.24 & 23.78 & 26.06 & 34.55 & 27.50 & 12.88 & 27.20 & 25.78 & 27.46 & 24.52 & 25.82 & 25.89 & 0.458 \\
MEDA~\cite{wan2025meda}        & 30\%  & 30\% & 0.80\,GB & 7.40 & 47.12 & 24.62 & 52.33 & 25.79 & 43.12 & 24.71 & 54.84 & 26.07 & 28.37 & 25.71 & 20.74 & 20.99 & 19.68 & 20.85 & 20.57 & 0.117 \\
HACK~\cite{qin2026head}        & 30\%  & 30\%  & 0.84\,GB & 7.40 & 27.59 & 27.25 & 28.02 & 27.40 & 21.70 & 26.32 & 27.36 & 27.99 & 8.62 & 27.76 & 28.12 & 30.11 & 26.97 & 28.23 & 28.36 & 0.658 \\
\ours HACK++ & 30\% & 20\%  & 0.61\,GB & 7.39 & 26.58 & 27.34 & 27.75 & 27.46 & 20.08 & 26.40 & 25.13 & 28.03 & 7.30 & 27.83 & 28.34  & 30.33   & 27.25  &  28.49 &28.60  & 0.680  \\
\bottomrule
\end{tabular}
}
\label{table1}
\end{table*}

\begin{figure*}[t]
  \centering
  \includegraphics[width=\linewidth]{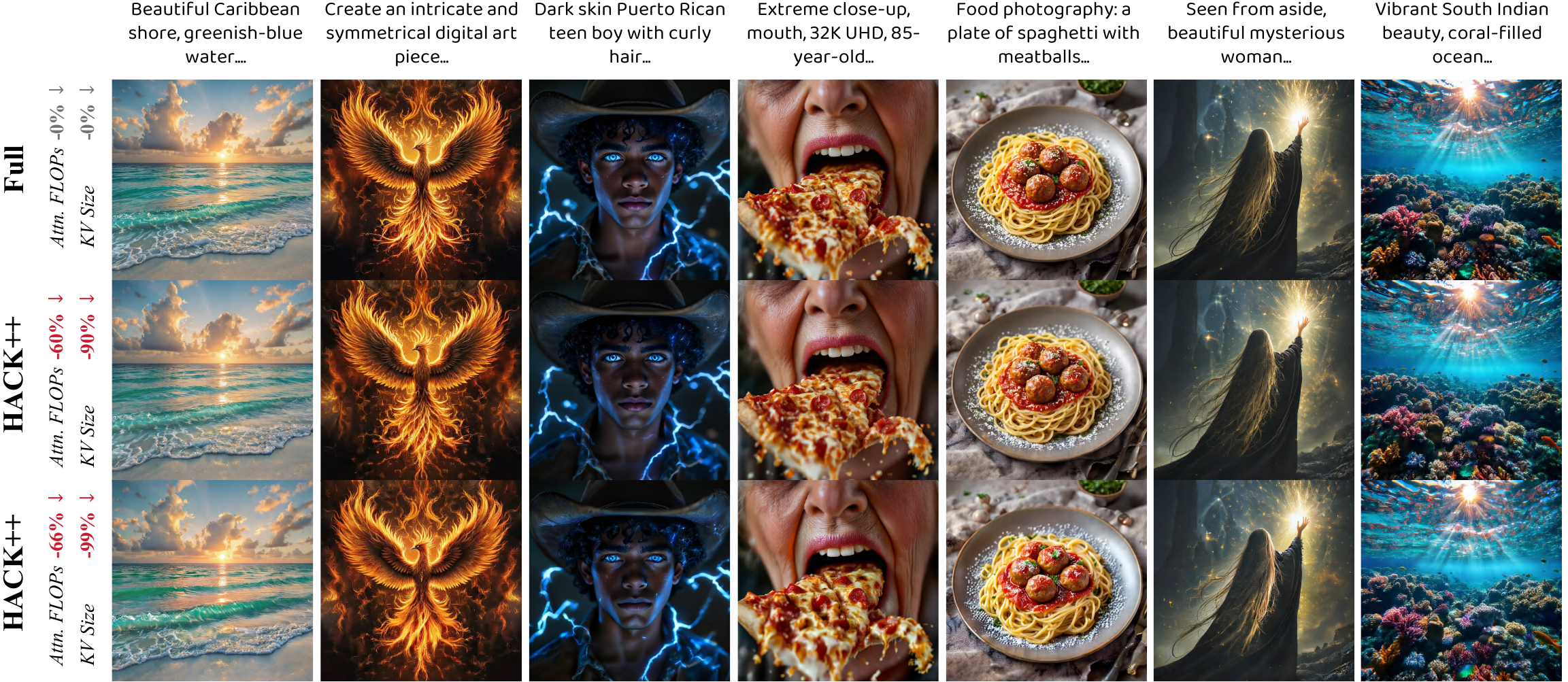}

\caption{Qualitative results of text-to-image generation by HACK++ on Infinity-8B~\cite{han2024infinity}. Rows show, from top to bottom, the Full baseline and HACK++ under two compression levels ($\eta_\text{a}=30\%, \eta_\text{c}=10\%$ and $\eta_\text{a}=30\%, \eta_\text{c}=1\%$).}
\label{8bcompare}
\end{figure*}
\section{Experiments}
 
\subsection{Experimental Settings}
 
\textbf{Evaluation Models.}
To demonstrate the broad applicability of HACK++, we conduct experiments on \textbf{seven} typical visual autoregressive
models spanning three paradigms: \textit{text-to-image generation}
(Infinity-2B, Infinity-8B~\cite{han2024infinity},
HART~\cite{tang2024hart}), \textit{class-conditional generation}
(VAR-d24, VAR-d30~\cite{tian2024visual}), and \textit{unified understanding
and generation} (VARGPT-v1.1~\cite{zhuang2025vargpt},
OneCAT-3B~\cite{Li2025OneCATDA}). For all models, we follow
the official evaluation protocols, including hyperparameter and
environment configurations, and fix the random seed across all runs
to ensure fair, reproducible comparisons.

\textbf{Evaluation Benchmarks and Metrics.}
For text-to-image generation, we adopt:
HPSv2.1~\cite{wu2023human} (HPS) and ImageReward~\cite{xu2023imagereward} (IR) for human preference alignment,
and MJHQ-30K~\cite{li2024playground} (overall and per-category FID and CLIP score) for distributional fidelity and text-image alignment.
For class-conditional generation, we report FID, Inception Score (IS), Precision, and Recall evaluated on ImageNet-1K~\cite{deng2009imagenet} using 50 samples per class.
For unified models, we evaluate on GenEval~\cite{ghosh2023geneval} and DPG~\cite{hu2024ella}, assessing compositional alignment and dense prompt-following.
We report KV Size (GB) and Attention FLOPs to quantify efficiency gains.
 
\textbf{Baseline Methods.}
We compare HACK++ against a wide range of existing KV cache compression methods, spanning three categories:
(1) \textit{Eviction-based}, including StreamingLLM~\cite{xiao2023efficient} (position-based), and H2O~\cite{zhang2024h2o}, SnapKV~\cite{li2024snapkv}, and CAKE~\cite{qin2025cake} (attention-based); 
(2) \textit{Merging-based}, including LOOK-M~\cite{wan2024look} and MEDA~\cite{wan2025meda}, which merge evicted KV pairs.
(3) \textit{VAR-specific methods}, including HACK~\cite{qin2026head} and ScaleKV~\cite{li2025scalekv}, which are specifically designed for the next-scale prediction paradigm.

\subsection{Implementation Details.}

\textbf{Offline Calibration Setting.}
For each evaluated model, we perform a one-time, offline calibration that jointly conducts head classification and per-head attention distribution estimation, using only \textit{50 calibration samples}. 
Specifically, we randomly sample prompts from ImageReward~\cite{xu2023imagereward} and HPSv2.1~\cite{wu2023human} benchmarks for Infinity-2B/8B, HART, VARGPT-v1.1, and OneCAT-3B; and class labels from ImageNet~\cite{deng2009imagenet} for VAR-d24 and VAR-d30. The entire calibration process completes \textit{within minutes} on a single GPU. To facilitate efficient inference, we statically reorder attention heads within each layer to group contextual and structural heads before deployment, eliminating runtime indexing overhead.

\textbf{Compression Setting.}
For each model, we adjust the attention budget ratio $\eta_a = B_a / T_K$ and cache budget ratio $\eta_c = B_c / T_K$ to accommodate varying backbones and target generation resolutions.
Given a target cache budget, HACK++ automatically derives a head-type-aware cache budget allocation from the calibrated attention distribution. For the attention budget, we adopt a uniform constraint across all heads, capping the KV pairs in each head's attention computation per scale.
For \textit{contextual heads}, we apply subset attention with a fixed subset size of $N_{\text{obs}} = 32$, uniformly sampled from full queries to estimate token importance, consistent across all models. For \textit{structural heads}, the scale-prior factors are precomputed offline and fixed throughout inference, incurring negligible online scoring cost.

\textbf{Model-Specific Configurations.}
We moderately adjust the contextual head ratio $\alpha$ and the sharpness coefficient $\tau$ to better align with each model's head count, variance distribution, and historical reliance. Specifically, we set the hyperparameter pair $(\alpha,\tau)$ to
$(0.2,1.0)$, $(0.3,2.0)$, and $(0.2,1.0)$ for Infinity-2B,
Infinity-8B, and HART, respectively. For unified models, we set
$(\alpha,\tau)$ to $(0.15,0.5)$ and $(0.3,1.5)$
for VARGPT and OneCAT, respectively. For class-conditional models,
we set $(\alpha,\tau)$ to $(0.15,0.5)$ and $(0.35,0.5)$ for
VAR-d24 and VAR-d30, respectively. We evaluate the robustness of
HACK++ to these hyperparameters in Sec.~\ref{sensi_study}.

\subsection{Main Results}
\textbf{Evaluation on Text-to-Image Generation.}
\textit{1) Quantitative Results.}
Tab.~\ref{table1} compares HACK++ with existing KV cache compression baselines on Infinity-2B, Infinity-8B, and HART. Across all three models, HACK++ achieves the best generation quality while also being the most efficient.
Existing baselines~\cite{xiao2023efficient,zhang2024h2o,li2024snapkv,qin2025cake,wan2024look,wan2025meda} designed for LLMs/VLMs transfer poorly to VAR: their unified importance estimation implicitly assumes the query-invariant, vertical attention of contextual heads and breaks down on structural heads, causing noticeable degradation in perceptual quality and human-preference scores under compression.
HACK++ surpasses HACK in both efficiency and generation quality,
owing to its better-designed compression with enhanced budget
allocation and importance estimation. 
Compared with ScaleKV, which targets extreme KV cache compression, HACK++ matches its cache reduction while additionally cutting attention FLOPs and delivering higher quality.  Notably, on Infinity-2B and -8B, HACK++
remains robust while keeping only $1\%$ of the KV cache, revealing
substantial redundancy in the cache of the vanilla next-scale process.
In several cases, HACK++ even surpasses the full-cache baseline---on
HART across most metrics---suggesting that redundant historical tokens
are not merely dispensable but can inject noise into next-scale
generation.

\begin{figure}[!t]
  \centering
  \includegraphics[width=1.0\linewidth]{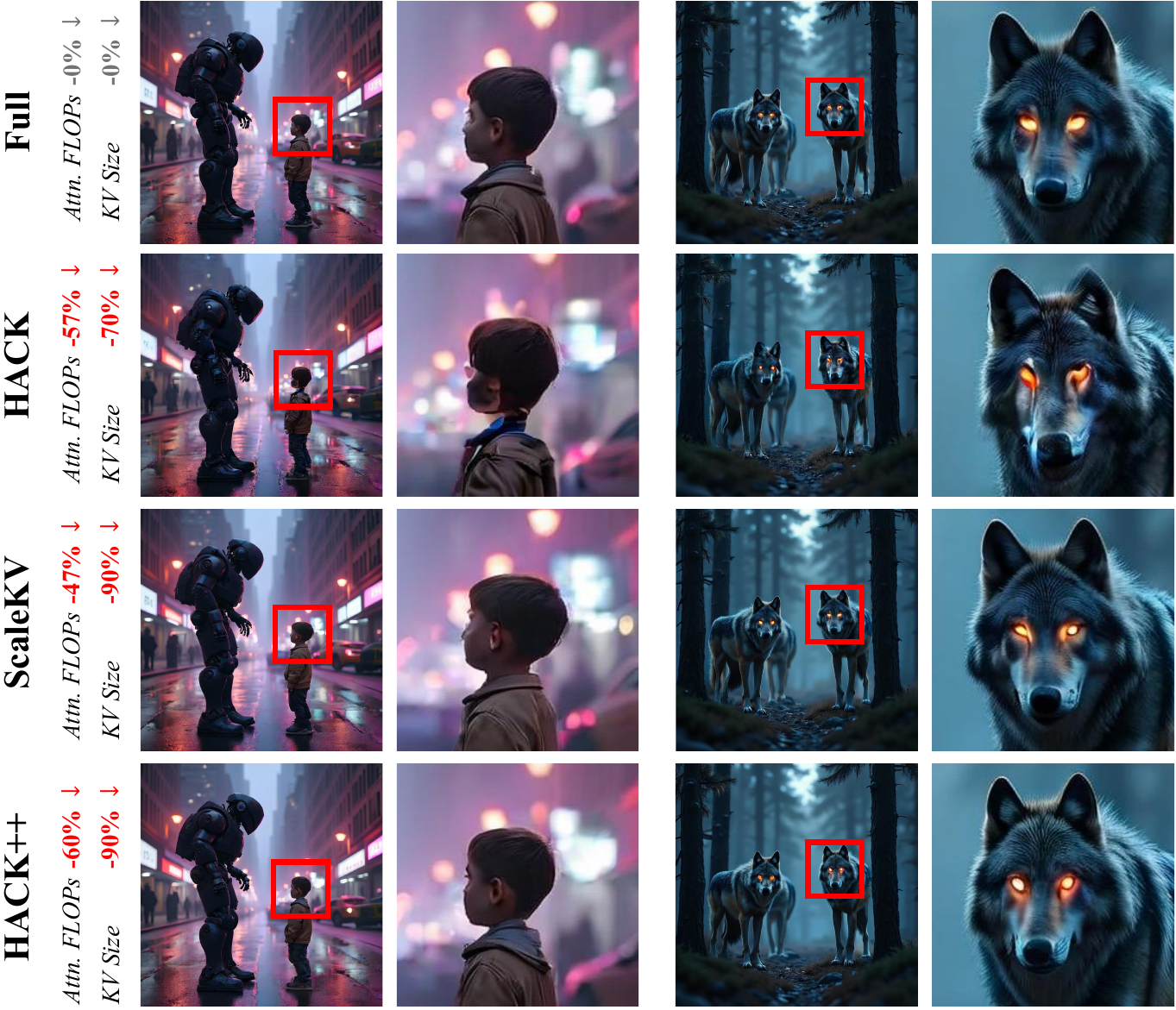}
\caption{Zoom-in comparison of HACK++ ($\eta_\text{a}=30\%,\,\eta_\text{c}=10\%$) with state-of-the-art HACK~\cite{qin2026head}($\eta_\text{a}=30\%,\,\eta_\text{c}=30\%$) and ScaleKV~\cite{li2025scalekv} ($\eta_\text{c}=10\%$) on Infinity-2B~\cite{han2024infinity}.}
  \label{zoomin}
\end{figure}

\textit{2) Qualitative Results.}
We further present qualitative examples of HACK++ on Infinity-8B in Fig.\,\ref{8bcompare}. Even under the extreme setting that reduces the KV cache by 99\%, HACK++ preserves visual fidelity, scene layout, object structure, and semantic consistency with the full-cache outputs across diverse prompts. We further provide a zoom-in comparison of HACK++ with HACK and ScaleKV on challenging fine-detail scenes (Fig.\,\ref{zoomin}). HACK occasionally distorts fine structures, as it aggressively compresses the attention computation, whereas ScaleKV preserves such details but reduces the attention FLOPs far less than HACK and HACK++. HACK++ enjoys the best of both worlds, delivering aggressive KV-cache and attention-FLOPs reduction while keeping intricate details sharp and aligned with Full.

\textbf{Evaluation on Class-Conditional Image Generation.}
\textit{1) Quantitative Results.}
We further evaluate HACK++ on class-conditional image generation with VAR-d30 and VAR-d24. As shown in Tab.\,\ref{table2}, this setting is highly sensitive to compression: the generated images have lower resolution and fewer tokens, leaving less redundancy to absorb
inaccurate KV compression. Baseline methods thus degrade sharply---for instance, LOOK-M raises the FID of VAR-d30 to $18.88$ at $\eta_a=\eta_c=30\%$, indicating a collapse in generation quality.
In contrast, both HACK and HACK++ remain robust under aggressive compression, owing to their head-aware, pattern-specific design for VAR models. Between the two, HACK++ achieves better FID and IS under
the same attention budget while reducing the KV cache more aggressively. These results show that the effectiveness of HACK++ extends beyond high-resolution text-to-image generation to the low-redundancy, class-conditional regime.

\begin{figure}[!t]
  \centering
  \includegraphics[width=0.8\linewidth]{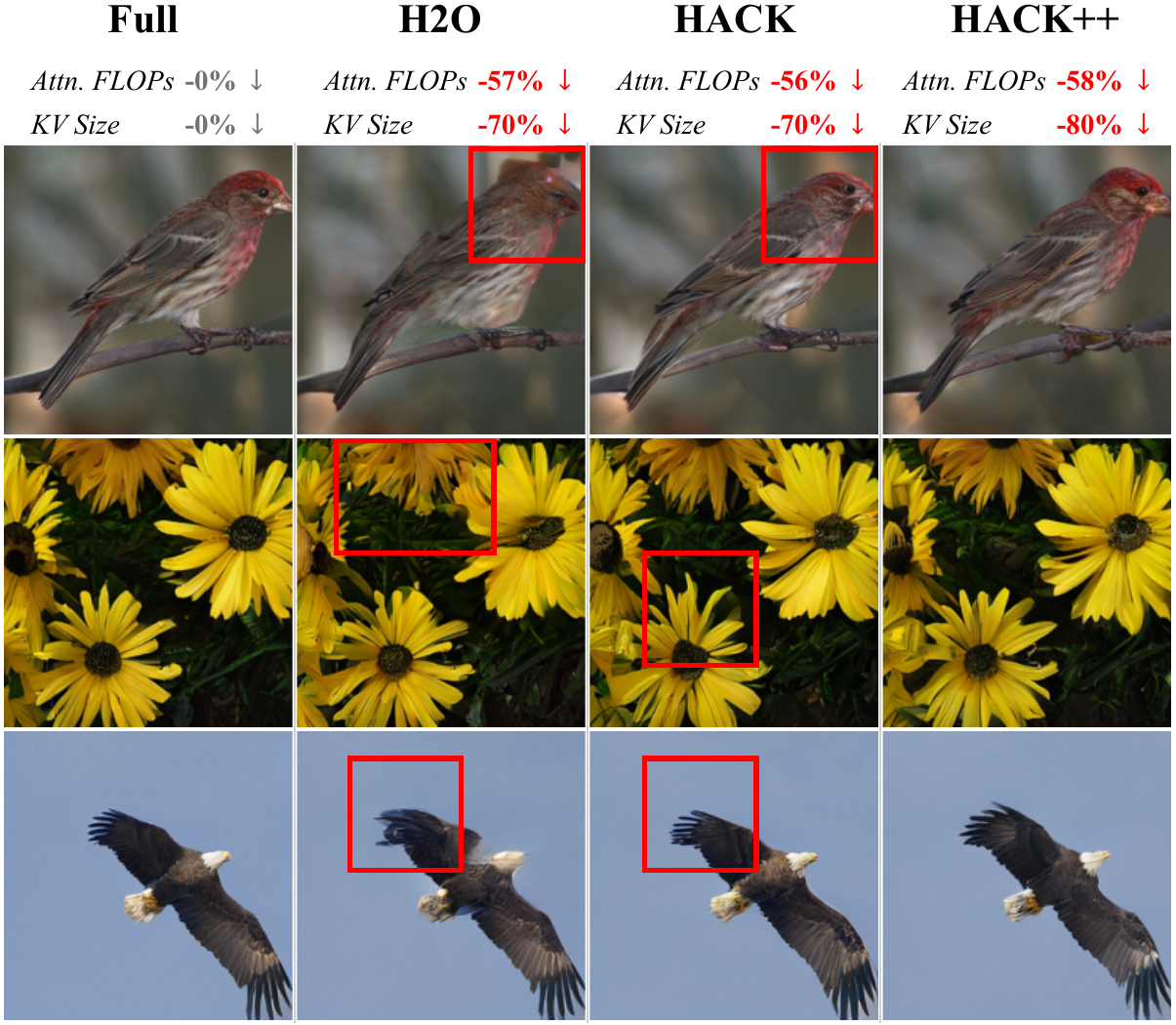}
\caption{Qualitative comparison of class-conditional image generation on VAR-d30~\cite{tian2024visual}. We compare HACK++ ($\eta_\text{a}=30\%,\,\eta_\text{c}=20\%$) with H2O and HACK ($\eta_\text{a}=\eta_\text{c}=30\%$).}
  \label{varquan}
\end{figure}

\textit{2) Qualitative Results.}
The qualitative results in Fig.\,\ref{varquan} are consistent with the quantitative findings. Under aggressive compression, H2O suffers from severe structural degradation, while HACK alleviates this but still leaves minor artifacts (highlighted in red). In contrast, HACK++ remains visually faithful to the Full baseline across all examples, preserving fine object details and texture continuity without noticeable degradation. Notably, HACK++ even attains this fidelity under more aggressive cache compression, demonstrating the effectiveness of the proposed design.

\begin{table*}[t]
\centering
\caption{Quantitative comparison of class-conditional image generation on ImageNet-1K~\cite{deng2009imagenet} using VAR-d30 and VAR-d24\cite{tian2024visual}. KV size and attention TFLOPs are
measured at batch size 32.
}
\label{tab:cls_cond}
\setlength{\tabcolsep}{4pt}
\begin{tabular}{l | cc | cccccc | cccccc}
\toprule
\multirow{2}{*}{Method}
 & \multirow{2}{*}{$\eta_\text{a}$}
 & \multirow{2}{*}{$\eta_\text{c}$}
 & \multicolumn{6}{c|}{\textit{VAR-d30}~\cite{tian2024visual}}
 & \multicolumn{6}{c}{\textit{VAR-d24}~\cite{tian2024visual}} \\
\cmidrule(lr){4-9}\cmidrule(lr){10-15}
 &&& KV Size$\downarrow$&TFLOPs$\downarrow$& FID$\downarrow$ & IS$\uparrow$ & Prec.$\uparrow$ & Rec.$\uparrow$
      & KV Size$\downarrow$&TFLOPs$\downarrow$& FID$\downarrow$ & IS$\uparrow$ & Prec.$\uparrow$ & Rec.$\uparrow$ \\
\midrule
Full & 100\% & 100\% & 14.01\,GB & 4.22& 1.96 & 302.23 & 0.81 & 0.60 & 8.96\,GB & 2.70 & 2.16 & 308.66 & 0.82 & 0.59  \\
\midrule
\multirow{2}{*}{Streaming~\cite{xiao2023efficient}}
 & 50\% & 50\% &  7.00\,GB & 2.73 & 2.36  & 281.30 & 0.79 & 0.61 & 4.48\,GB & 1.75 & 2.42 & 284.56 & 0.80 & 0.60 \\
 & 30\% & 30\% &  4.20\,GB & 1.80  & 4.84  & 228.43 & 0.74 & 0.62 & 2.69\,GB & 1.15 & 5.19 & 227.64 & 0.74 & 0.61 \\
\midrule
\multirow{2}{*}{H2O~\cite{zhang2024h2o}}
 & 50\% & 50\% &  7.00\,GB & 2.73 & 3.04  & 262.68 & 0.77 & 0.62 & 4.48\,GB & 1.75 & 2.64 & 273.55 & 0.78 & 0.61  \\
 & 30\% & 30\% & 4.20\,GB & 1.80& 8.81  & 182.84 & 0.68 & 0.62 & 2.69\,GB & 1.15 & 8.14 & 187.47 & 0.68 & 0.63 \\
\midrule
\multirow{2}{*}{SnapKV~\cite{li2024snapkv}}
 & 50\% & 50\% & 7.00\,GB & 2.73  & 3.09  & 261.63 & 0.77 & 0.62 & 4.48\,GB & 1.75 & 2.79 & 270.32 & 0.77 & 0.62 \\
 & 30\% & 30\% & 4.20\,GB & 1.80 & 7.31  & 196.62 & 0.70 & 0.62 & 2.69\,GB & 1.15 & 7.15 & 198.83 & 0.69 & 0.63 \\
\midrule
\multirow{2}{*}{LOOK-M~\cite{wan2024look}}
 & 50\% & 50\% &  7.00\,GB & 2.73  & 6.89  & 203.47 & 0.70 & 0.62 & 4.48\,GB & 1.75 & 6.01 & 212.47 & 0.70 & 0.63 \\
 & 30\% & 30\% & 4.20\,GB & 1.80 & 18.88 & 116.70 & 0.58 & 0.63 & 2.69\,GB & 1.15 & 18.28 & 120.18 & 0.57 & 0.63 \\
\midrule
\multirow{2}{*}{HACK~\cite{qin2026head}}
 & 50\% & 50\% & 7.00\,GB & 2.73 & 2.06  & 293.60 & 0.80 & 0.61 & 4.48\,GB & 1.76 & 2.22 & 294.87 & 0.80 & 0.60 \\
 & 30\% & 30\% & 4.20\,GB & 1.86 & 2.78  & 268.69 & 0.78 & 0.62 & 2.69\,GB & 1.19 & 4.02 & 243.50 & 0.75 & 0.62 \\
\midrule
\multirow{2}{*}{HACK++}

& \cellcolor{lightgray}50\% & \cellcolor{lightgray}30\% & \cellcolor{lightgray}4.20\,GB & \cellcolor{lightgray}2.59 & \cellcolor{lightgray}2.05 & \cellcolor{lightgray}302.49 & \cellcolor{lightgray}0.81 & \cellcolor{lightgray}0.59 & \cellcolor{lightgray}2.69\,GB & \cellcolor{lightgray}1.66 & \cellcolor{lightgray}2.21 & \cellcolor{lightgray}300.78 & \cellcolor{lightgray}0.82 & \cellcolor{lightgray}0.59 \\
& \cellcolor{lightgray}30\% & \cellcolor{lightgray}20\% & \cellcolor{lightgray}2.80\,GB & \cellcolor{lightgray}1.78 & \cellcolor{lightgray}2.48 & \cellcolor{lightgray}282.67 & \cellcolor{lightgray}0.79 & \cellcolor{lightgray}0.59 & \cellcolor{lightgray}1.79\,GB & \cellcolor{lightgray}1.14 & \cellcolor{lightgray}3.17 & \cellcolor{lightgray}262.77 & \cellcolor{lightgray}0.80 & \cellcolor{lightgray}0.59 \\
\bottomrule
\end{tabular}
\label{table2}
\end{table*}

\begin{table*}[t]
\centering
\caption{Generalization to unified understanding-and-generation models
VARGPT-v1.1~\cite{zhuang2025vargpt} and OneCAT-3B~\cite{Li2025OneCATDA}
on the GenEval~\cite{ghosh2023geneval} and DPG-Bench~\cite{hu2024ella}
benchmarks. KV size and attention FLOPs are measured at batch size 1.}
\label{tab:unified}
\setlength{\tabcolsep}{2pt}
\resizebox{\textwidth}{!}{
\begin{tabular}{l | l cc cc | cccccc | cccc}
\toprule
\multirow{2}{*}{Model}
& \multirow{2}{*}{Method}
& \multirow{2}{*}{$\eta_\text{a}$}
& \multirow{2}{*}{$\eta_\text{c}$}
& \multirow{2}{*}{\makecell{KV \\Size$\downarrow$}}
& \multirow{2}{*}{\makecell{Attn.\\TFLOPs$\downarrow$}}
& \multicolumn{6}{c|}{GenEval $\uparrow$}
& \multicolumn{4}{c}{DPG $\uparrow$} \\
\cmidrule(lr){7-12} \cmidrule(lr){13-16}
& & & & &
& Single Obj. & Two Obj. & Position & Color & Attri. & Overall
& Entity & Relation & Attribute & Overall \\
\midrule
\multirow{3}{*}{\textit{VARGPT-v1.1}\cite{zhuang2025vargpt}}
& Full
& 100\% & 100\% & 1.23\,GB & 2.09
& 0.94 & 0.46 & 0.08 & 0.76 & 0.10 & 0.46
& 82.78 & 88.86 & 82.43 & 75.28 \\
& HACK~\cite{qin2026head} 
& 30\% & 30\% & 0.37\,GB & 0.89
& 0.96 & 0.46 & 0.08 & 0.74 & 0.13 & 0.45
& 83.09 & 89.90 & 82.49 & 75.20 \\
& \cellcolor{lightgray}HACK++
& \cellcolor{lightgray}30\% & \cellcolor{lightgray}10\% 
& \cellcolor{lightgray}0.12\,GB  & \cellcolor{lightgray}0.87
& \cellcolor{lightgray}0.96 & \cellcolor{lightgray}0.43 & \cellcolor{lightgray}0.07 & \cellcolor{lightgray}0.76 
& \cellcolor{lightgray}0.10  & \cellcolor{lightgray}0.46
& \cellcolor{lightgray}83.86 & \cellcolor{lightgray}89.83 & \cellcolor{lightgray}82.15 & \cellcolor{lightgray}75.60 \\
\midrule

\multirow{3}{*}{\textit{OneCAT-3B}~\cite{Li2025OneCATDA}}
& Full
& 100\% & 100\% & 0.72\,GB & 40.66
& 0.99 & 0.94 & 0.84 & 0.95 & 0.76 & 0.88
& 89.71 & 93.73 & 85.75 & 83.23 \\
& HACK~\cite{qin2026head} 
& 30\% & 30\% &0.26\,GB  & 18.58
& 0.99 & 0.94 & 0.83 & 0.92 & 0.72 & 0.86
& 89.55 & 93.85 & 85.99 & 83.73 \\
& \cellcolor{lightgray}HACK++
& \cellcolor{lightgray}30\% & \cellcolor{lightgray}10\%
& \cellcolor{lightgray}0.07\,GB & \cellcolor{lightgray}16.55
& \cellcolor{lightgray}1.00 & \cellcolor{lightgray}0.94 & \cellcolor{lightgray}0.85 & \cellcolor{lightgray}0.93
& \cellcolor{lightgray}0.74 & \cellcolor{lightgray}0.87
& \cellcolor{lightgray}89.76 & \cellcolor{lightgray}93.73 & \cellcolor{lightgray}85.75 & \cellcolor{lightgray}83.23 \\
\bottomrule
\end{tabular}
}
\end{table*}

 \begin{figure}[!t]
  \centering
  \includegraphics[width=1.0\linewidth]{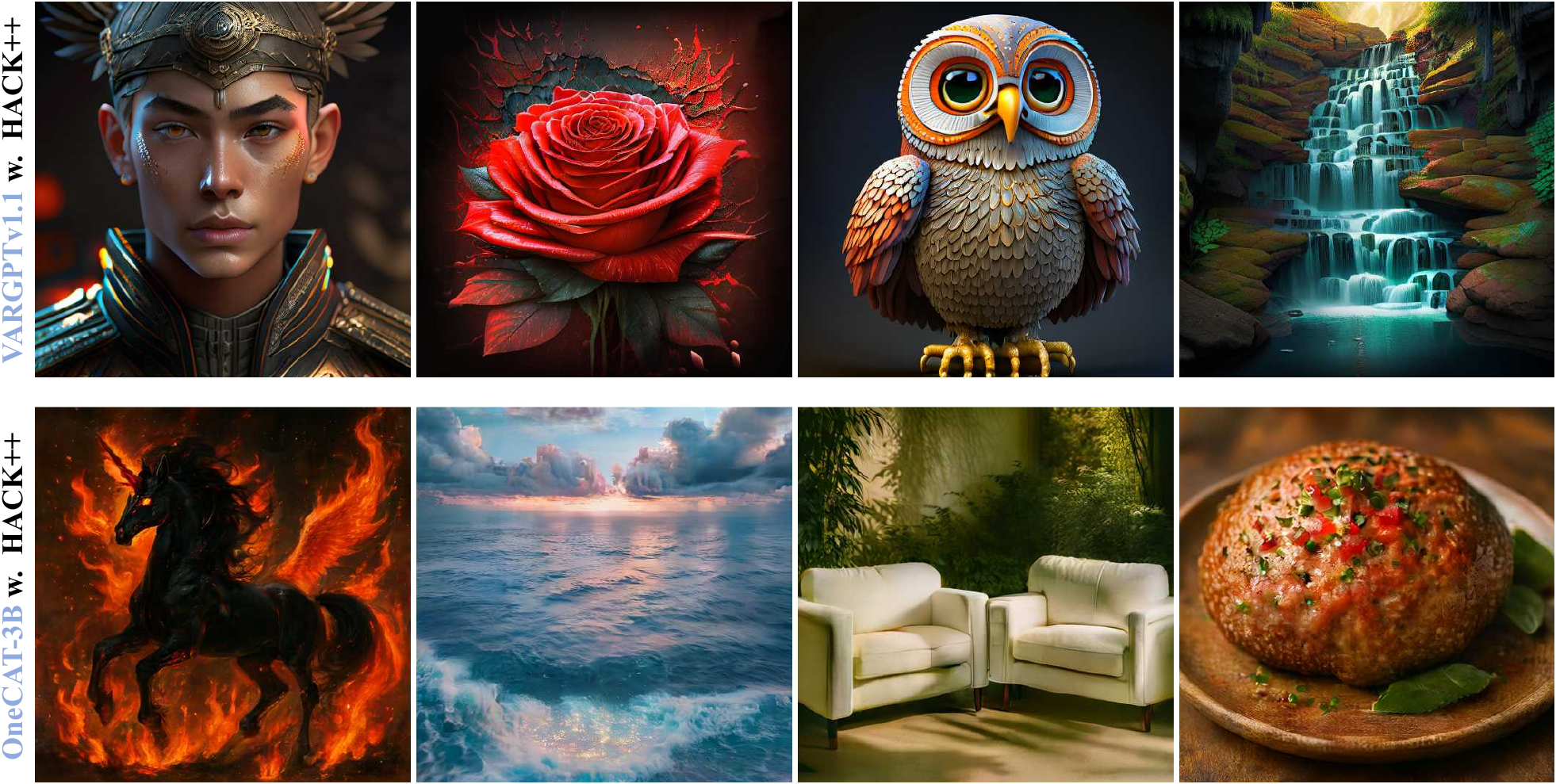}
\caption{Qualitative results for VARGPT-v1.1~\cite{zhuang2025vargpt} and OneCAT-3B~\cite{Li2025OneCATDA} using HACK++ ($\eta_\text{a}=30\%,\,\eta_\text{c}=10\%$).}
  \label{unified}
\end{figure}
\textbf{Generalization to Unified VAR Models.}
We further extend the evaluation to unified
understanding-and-generation VAR models, VARGPT-v1.1 and OneCAT-3B.
As shown in Tab.\,\ref{tab:unified}, both HACK and HACK++ stay within
$0.01$ of the full-cache baseline on GenEval overall and match or
slightly exceed it on DPG, confirming that the contextual--
structural dichotomy of VAR attention---and thus our pattern-specific
compression---carries over to the generation branch of unified models.
The advantage of HACK++ lies in efficiency: under the same attention
budget ($\eta_a=30\%$) as HACK, it pushes the cache ratio from
$\eta_c=30\%$ to $10\%$ without metric loss, reducing the KV cache
from $0.37$ to $0.12$\,GB on VARGPT-v1.1 and from $0.26$ to
$0.07$\,GB on OneCAT-3B---about a $3\times$ further reduction. We further visualize the generation results of these unified models with HACK++ in Fig.\,\ref{unified}, demonstrating its effectiveness on unified models.
\begin{table*}[t]
\centering
\caption{Efficiency comparison across VAR models under standard
attention. Measured on a single $96$\,GB
NVIDIA RTX PRO 6000 GPU with batch size $1$.
``Com.\,O.'' denotes the additional overhead introduced by the
compression itself; ``Mem.\,Reduction'' and ``Thr.\,Speedup'' are
reported relative to the full-cache baseline of each model.}
\label{tab:efficiency}
\setlength{\tabcolsep}{8pt}
\renewcommand{\arraystretch}{1.05}
\begin{tabular}{l | l | cccccc}
\toprule
Model & Method
 & Memory $\downarrow$ & Mem. Saving $\uparrow$
& Com O. $\downarrow$ & Latency $\downarrow$ & Throughput  $\uparrow$ & Thr. Speedup $\uparrow$ \\
\midrule
\multirow{4}{*}{\textit{Infinity-2B}~\cite{han2024infinity}}
& Full                &28.28\,GB  &1.00$\times$  &--  &2.35\,s  &0.43\,it/s  &1.00$\times$  \\
& HACK~\cite{qin2026head}                  &14.50\,GB  &1.95$\times$ &0.17\,s &1.71\,s   &0.59\,it/s &1.37$\times$  \\
& ScaleKV~\cite{li2025scalekv}                 & 21.37\,GB    & 1.32$\times$           & 0.37\,s   & 2.09\,s   & 0.48\,it/s      & 1.12$\times$     \\
& HACK++                  &12.88\,GB  &2.20$\times$    &0.08\,s           &1.54\,s   & 0.65\,it/s   & 1.51$\times$           \\
\midrule
 \multirow{4}{*}{\textit{Infinity-8B}~\cite{han2024infinity}}
& Full                    &59.00\,GB  &1.00$\times$  &--  &4.88\,s  &0.21\,it/s  &1.00$\times$ \\
& HACK~\cite{qin2026head}                    &32.34\,GB  &1.82$\times$  &0.33\,s & 3.45\,s &0.29\,it/s  &1.38$\times$  \\
& ScaleKV~\cite{li2025scalekv}                  &43.88\,GB  &1.34$\times$    &0.82\,s           &4.29\,s   &0.23\,it/s   &1.10$\times$           \\
& HACK++                  &28.89\,GB  & 2.04$\times$    & 0.14\,s          &3.11\,s  & 0.32\,it/s   & 1.52$\times$           \\
\midrule
\multirow{3}{*}{\textit{HART}~\cite{tang2024hart}}
& Full                    &31.02\,GB  &1.00$\times$  &--  &1.60\,s  &0.63\,it/s  &1.00$\times$  \\
& HACK~\cite{qin2026head}           & 20.67\,GB & 1.50$\times$	& 0.19\,s& 1.16\,s & 0.86\,it/s &1.37$\times$  \\
& HACK++                  & 17.66\,GB & 1.76$\times$    & 0.03\,s           & 0.99\,s   & 1.01\,it/s   & 1.60$\times$           \\
 
\bottomrule
\end{tabular}
\end{table*}
\subsection{Efficiency Analysis}
We evaluate the efficiency of HACK++ by comparing memory consumption
and inference latency across VAR models under a standard attention
implementation. Tab.\,\ref{tab:efficiency} reports the results, including the
additional compression overhead (Com.\,O) incurred by each method. Vanilla VAR models suffer from intensive attention computation
and a cumulative KV cache; HACK++ substantially reduces both, cutting
the memory footprint by $2.04\times$ and accelerating inference by
$1.52\times$ on Infinity-8B. The decoupled pipeline lets HACK++
dominate both prior methods on their respective strengths. Against
HACK, which couples the two budgets, HACK++ compresses the cache more
aggressively (e.g., $28.89$\,GB vs. $32.34$\,GB on Infinity-8B, a
$2.04\times$ vs. $1.82\times$ reduction) while \emph{lowering} the
compression overhead from $0.33$\,s to $0.14$\,s. Against ScaleKV, which focuses on the cache reduction, HACK++ additionally throttles attention
computation, reaching a $1.52\times$ throughput speedup versus
ScaleKV's $1.10\times$ at a smaller memory footprint
($28.89$\,GB vs. $43.88$\,GB). HACK++ thus attains the best overall
speed--memory trade-off, and its compression is near-free, accounting
for about $5\%$ of total latency on all three models.

\begin{figure}[!t]
  \centering
  \includegraphics[width=1.0\linewidth]{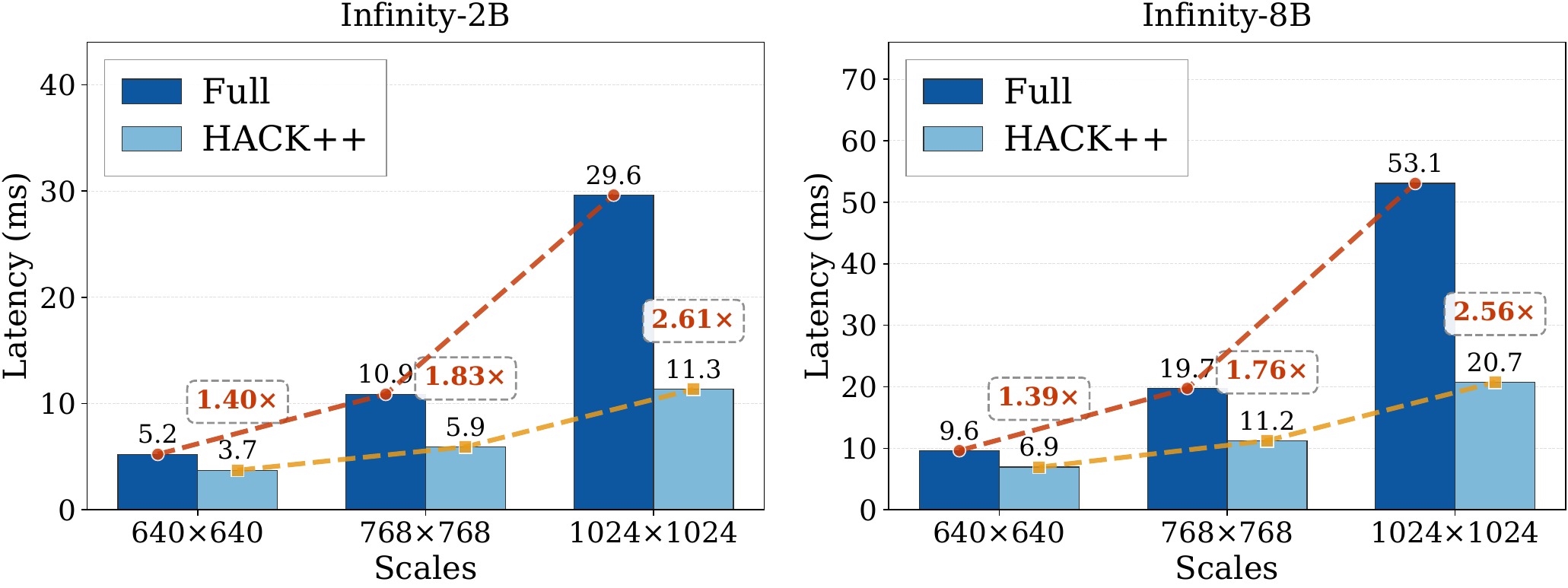}
  \caption{Efficiency profiling of average latency on attention module for different scales on Infinity models, evaluated on a single
$96$\,GB NVIDIA RTX PRO 6000 GPU. }
  \label{attn_profile}
\end{figure}

\begin{figure}[!t]
  \centering
  \includegraphics[width=1.0\linewidth]{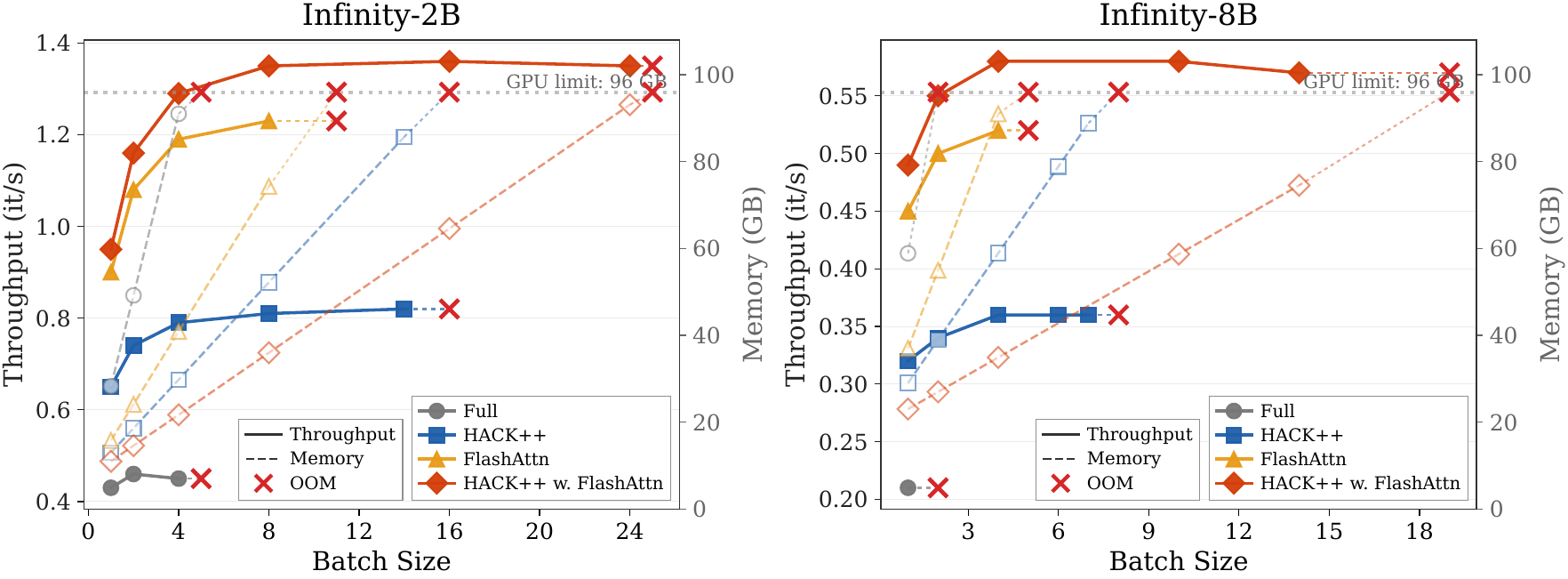}
  \caption{Throughput and memory comparison between the full cache
and HACK++ on Infinity models, evaluated on a single $96$\,GB
NVIDIA RTX PRO 6000 GPU.}
  \label{flashattn_profile}
\end{figure}

HACK++ also scales gracefully to higher resolutions. As shown in
Fig.\,\ref{attn_profile}, the per-scale attention latency of full
attention grows steeply with resolution, whereas HACK++ holds this
growth close to linear, yielding a $2.6\times$ speedup at
$1024\times1024$. Finally, HACK++ is orthogonal to acceleration
frameworks such as FlashAttention~\cite{dao2023flashattention} and
compounds with them. 
As shown in Fig.\,\ref{flashattn_profile}, on a $96$\,GB NVIDIA RTX
PRO 6000 GPU the vanilla Infinity-8B exhausts memory beyond a batch
size of $1$, and even its FlashAttention variant fails beyond $4$. By
shortening the KV cache itself, HACK++ scales to a batch size of $7$
under standard attention and to $18$ when combined with FlashAttention,
raising peak throughput from $0.21$\,it/s to $0.58$\,it/s, a $2.76\times$
gain over the full-cache baseline. These results confirm the robust
efficiency of HACK++ and its compatibility with existing
optimizations, making it well suited for resource-constrained deployment.

\subsection{Ablation Study}
\textbf{Ablation on Key Components.}
Table\,\ref{ablation_key_components} examines the two key components of HACK++: head-aware compression via pattern-specific importance estimation and adaptive cache budget allocation. Removing both components results in significant performance drops, as a unified compression strategy and a uniform budget allocation fail to accommodate the divergent attention patterns and heterogeneous historical reliance across heads. Pattern-specific importance estimation improves performance by retaining the right tokens for each head type, while adaptive budget allocation improves performance by distributing the cache budget according to each head group's actual demand. The full HACK++, combining both, consistently achieves the best performance across all metrics, demonstrating their complementary benefits.


 \begin{table}[t]
\centering
\caption{Component ablation of HACK++ on Infinity-2B ($\eta_\text{a}=30\%, \eta_\text{c}=1\%$) and HART ($\eta_\text{a}=30\%, \eta_\text{c}=20\%$).
``Adapt. Budget'' denotes the adaptive budget allocation;
``Pattern Comp.'' denotes the pattern-specific compression.}
\label{tab:ablation_component}
\setlength{\tabcolsep}{8pt}
\renewcommand{\arraystretch}{1.05}
\begin{tabular}{cc | cc | cc}
\toprule
\makecell{Adapt.\\Budget}
& \makecell{Pattern\\Comp.}
& \multicolumn{2}{c|}{\textit{Infinity-2B}~\cite{han2024infinity}}
& \multicolumn{2}{c}{\textit{HART}~\cite{tang2024hart}}\\
\cmidrule(lr){3-4} \cmidrule(lr){5-6}
& &  IR $\uparrow$ & HPS $\uparrow$ & IR $\uparrow$ & HPS $\uparrow$ \\
\midrule
\ding{55} & \ding{55}  & 0.892& 29.68 &  0.531 & 26.53 \\
\ding{51} & \ding{55} & 0.914 & 29.96 &   0.558 & 26.68 \\
\ding{55} & \ding{51}  & 0.897  &  29.89  & 0.572 & 27.37 \\
\ours\ding{51} & \ding{51}  & \textbf{0.934}   & \textbf{30.25}       & \textbf{0.680} & \textbf{28.60} \\
\bottomrule
\end{tabular}
\label{ablation_key_components}
\end{table}

\textbf{Ablation on Pattern-Specific Compression.}
\label{sec:ablation_pattern}
We ablate the internal design of $f_C$ and $f_S$, then verify the
necessity of head-aware strategy assignment.

\paragraph{Importance estimation for contextual heads}
We first analyze $f_C$, which scores contextual-head importance from
attention computed over a uniformly sampled query subset.
Tab.\,\ref{tab:ablation_contextual_scoring} (top) compares three
sampling strategies at a fixed $N_{\text{obs}}{=}32$. Owing to the
column-wise stability of contextual attention
(Sec.\,\ref{sec:patterns}), the three strategies perform almost
identically; uniform sampling is marginally best on Infinity-2B and
on par with the others on HART, so we adopt it for its balanced
spatial coverage. The bottom block varies the subset size: accuracy
saturates by $N_{\text{obs}}{=}32$, and larger subsets yield no gain,
with even a slight drop on HART, at higher scoring cost. We therefore
set $N_{\text{obs}}{=}32$.

\paragraph{Importance estimation for structural heads}
We next ablate $f_S$ (Tab.\,\ref{ab_sh}). Plain query-subset attention
is inadequate for structural heads: as analyzed in
Sec.\,\ref{sec:scoring_structural}, sampled queries cannot locate
position-sensitive structural anchors and degenerate into near-uniform
within-scale sampling. HACK augments this with a fixed recent-scale
heuristic, which helps but neither adapts to per-head/layer scale
preferences (Sec.\,\ref{sec:heterogeneity}) nor selects informative
anchors within a scale. HACK++ instead derives calibrated, per-head
scale-prior factors. The scale prior and the value norm are
complementary: the former selects the relevant source scales but cannot
discriminate tokens within a scale, while the latter pinpoints
high-impact anchors within a scale but is blind to which scales the head
prefers. Using either alone is suboptimal; combining them yields the
best performance.

\paragraph{Head-aware strategy assignment}
Tab.\,\ref{tab:ablation_assignment} verifies the need to match each
head type to its own strategy. Applying a single strategy to all
heads, either $f_C$ or $f_S$, underperforms the head-aware design:
$f_C$ collapses on structural heads, whose sampled queries degenerate
into uniform within-scale sampling
(Sec.\,\ref{sec:scoring_structural}), while $f_S$ lacks the
query-driven semantic selectivity that contextual heads require.
Swapping the assignment ($f_C\!\leftrightarrow\!f_S$) gives the worst
result of all, confirming that the two head types not only need
different strategies but need their specifically matched ones.

\begin{table}[t]
\centering
\caption{Ablation on contextual-head importance estimation ($f_C$), using Infinity-2B ($\eta_\text{a}=30\%, \eta_\text{c}=1\%$) and HART ($\eta_\text{a}=30\%, \eta_\text{c}=20\%$). Top: query sampling strategy under a fixed $N_{\text{obs}}{=}32$. Bottom: effect of the subset size $N_{\text{obs}}$.}
\label{tab:ablation_contextual_scoring}
\setlength{\tabcolsep}{8pt}
\begin{tabular}{l | cc | cc}
\toprule
& \multicolumn{2}{c|}{\textit{Infinity-2B}~\cite{han2024infinity}}
& \multicolumn{2}{c}{\textit{HART}~\cite{tang2024hart}} \\
\cmidrule(lr){2-3} \cmidrule(lr){4-5}
& IR $\uparrow$ & HPS $\uparrow$ & IR $\uparrow$ & HPS $\uparrow$\\
\midrule
\multicolumn{5}{c}{Query sampling strategy ($N_{\text{obs}}{=}32$)}\\
\midrule
Initial & 0.930 & 30.18 & 0.679 & 28.60 \\
Last    & 0.925 & 30.17 & 0.683 & 28.61 \\
\ours Uniform & 0.934 & 30.25 & 0.680 & 28.60 \\
\midrule
\multicolumn{5}{c}{Subset size $N_{\text{obs}}$}\\
\midrule
8   & 0.920 & 30.23 & 0.674 & 28.60 \\
16  & 0.932 & 30.19 & 0.680 & 28.60 \\
\ours32  & 0.934 & 30.25 & 0.680 & 28.60 \\
64  & 0.934 & 30.26 & 0.678 & 28.60 \\
128 & 0.933 & 30.27 & 0.674 & 28.59 \\
Full & 0.934 & 30.27 & 0.676 & 28.60 \\
\bottomrule
\end{tabular}
\end{table}

\begin{table}[t]
\centering
\caption{Ablation on structural-head importance estimation ($f_S$), using Infinity-2B ($\eta_\text{a}=30\%, \eta_\text{c}=1\%$) and HART ($\eta_\text{a}=30\%, \eta_\text{c}=20\%$).}
\setlength{\tabcolsep}{6pt}
\renewcommand{\arraystretch}{1.05}
\begin{tabular}{l | cc | cc}
\toprule
\multirow{2}{*}{Strategy}
& \multicolumn{2}{c|}{\textit{Infinity-2B}~\cite{han2024infinity}}
& \multicolumn{2}{c}{\textit{HART}~\cite{tang2024hart}} \\
\cmidrule(lr){2-3} \cmidrule(lr){4-5}
& IR $\uparrow$ & HPS $\uparrow$ & IR $\uparrow$ & HPS $\uparrow$\\
\midrule
Query-subset attention      & 0.914 & 29.96 & 0.558 & 26.68 \\
Query-subset + recent (HACK)& 0.917 & 30.01 & 0.607 & 27.84 \\
Value norm only             & 0.918 & 30.18 & 0.498 & 26.48 \\
Scale-prior factor only     & 0.913 & 30.06 & 0.636 & 28.15 \\
\midrule
\ours Scale-prior factor + value norm & \textbf{0.934} & \textbf{30.25} & \textbf{0.680} & \textbf{28.60} \\
\bottomrule
\end{tabular}
\label{ab_sh}
\end{table}

\begin{table}[t]
\centering
\caption{Ablation on head-aware strategy assignment, using Infinity-2B ($\eta_\text{a}=30\%, \eta_\text{c}=1\%$) and HART ($\eta_\text{a}=30\%, \eta_\text{c}=20\%$). $f_C \!\leftrightarrow\! f_S$ swaps the strategy assigned to each head type.}
\label{tab:ablation_assignment}
\setlength{\tabcolsep}{8pt}
\renewcommand{\arraystretch}{1.05}
\begin{tabular}{l | cc | cc}
\toprule
\multirow{2}{*}{Strategy}
& \multicolumn{2}{c|}{\textit{Infinity-2B}~\cite{han2024infinity}}
& \multicolumn{2}{c}{\textit{HART}~\cite{tang2024hart}} \\
\cmidrule(lr){2-3} \cmidrule(lr){4-5}
& IR $\uparrow$ & HPS $\uparrow$ & IR $\uparrow$ & HPS $\uparrow$\\
\midrule
only $f_C$                    & 0.914 & 29.96  & 0.558 & 26.68 \\
only $f_S$         & 0.928 & 30.13  & 0.674&   28.49\\
$f_C \leftrightarrow f_S$ (Swapped)& 0.910 & 29.89  & 0.489 & 26.06  \\

\midrule
\ours Full HACK++ ($f_C + f_S$) & \textbf{0.934}   & \textbf{30.25}        & \textbf{0.680} & \textbf{28.60} \\
\bottomrule
\end{tabular}
\end{table}

\textbf{Ablation on Budget Allocation.}
\label{sec:ablation_budget}
We investigate the effectiveness of HACK++'s adaptive 
budget allocation by ablating along the three axes of 
heterogeneity identified in Sec.\,\ref{sec:heterogeneity}: 
head type, layer, and generation step. 
Tab.\,\ref{tab:ablation_budget} compares HACK++ against 
uniform allocation and three ablated variants, each 
removing one or more axes of adaptivity from HACK++.
All ablated variants outperform uniform allocation, 
confirming that each axis of adaptivity contributes 
meaningfully to allocation quality. However, no partial 
adaptivity recovers the full performance: removing 
head-type adaptivity fails to exploit the divergent 
compression sensitivity between contextual and 
structural heads; removing layer adaptivity ignores the 
markedly different historical reliance across layers; 
and removing step adaptivity mismatches the actual 
reliance demands at earlier generation steps, since the 
reliance pattern shifts as generation proceeds and 
cannot be faithfully represented by any single-step or 
step-averaged allocation. HACK++ achieves the best 
performance on both tasks by adapting simultaneously 
across all three axes, validating that full three-axis 
adaptivity is necessary.

\textbf{Ablation on Decoupled Attention and Cache Compression.}
\label{sec:decouple_ablation}
We ablate the two compression stages of HACK++ independently in
Tab.~\ref{tab:decouple_ablation}; the two budgets target orthogonal bottlenecks.
Compressing only the attention stage substantially reduces compute (TFLOPs) while
leaving the KV cache untouched, with negligible quality change. Compressing only the
cache stage instead shrinks KV memory by an order of magnitude and, rather than
hurting quality, slightly improves it, suggesting that the evicted tokens are largely
redundant. Because the two stages act on different resources, enabling both is
complementary: HACK++ attains the lowest compute and memory simultaneously while
matching the uncompressed quality. This validates our decoupled design—separating the
attention budget $\eta_\text{a}$ from the cache budget $\eta_\text{c}$ lets each
bottleneck be compressed at its own ratio ($\eta_\text{c}\!\ll\!\eta_\text{a}$),
which a single shared budget cannot.

\begin{table}[t]
\centering
\caption{Ablation on cache budget allocation along the 
three axes of heterogeneity (head type, layer, 
generation step), using Infinity-2B ($\eta_\text{a}=30\%, \eta_\text{c}=1\%$) and HART ($\eta_\text{a}=30\%, \eta_\text{c}=20\%$).}
\label{tab:ablation_budget}
\setlength{\tabcolsep}{8pt}
\renewcommand{\arraystretch}{1.05}
\begin{tabular}{ccc | cc | cc}
\toprule
\multicolumn{3}{c|}{Adaptive axes}
& \multicolumn{2}{c|}{\textit{Infinity-2B}~\cite{han2024infinity}}
& \multicolumn{2}{c}{\textit{HART}~\cite{tang2024hart}}\\
\cmidrule(lr){1-3} \cmidrule(lr){4-5} \cmidrule(lr){6-7}
Type & Layer & Step
& IR $\uparrow$ & HPS $\uparrow$ 
& IR $\uparrow$ & HPS $\uparrow$ \\
\midrule
\ding{55} & \ding{55} & \ding{55} 
& 0.897  &  29.89  & 0.572 & 27.37  \\
\ding{55} & \ding{51} & \ding{51} 
&0.920  & 30.18  & 0.669 & 28.35 \\

\ding{51} & \ding{55} & \ding{51} 
& 0.913   & 30.06   & 0.653  &  28.13   \\

\ding{51} & \ding{51} & \ding{55} 
& 0.931   &  30.16   & 0.675  & 28.51    \\

\midrule
\rowcolor{gray!10}
\ding{51} & \ding{51} & \ding{51} 
& \textbf{0.934} & \textbf{30.25} & \textbf{0.680} & \textbf{28.60} \\
\bottomrule
\end{tabular}
\end{table}

\begin{table}[t]
\centering
\caption{Ablation on decoupled compression on Infinity-2B. We toggle the pre-attention (attn.) and post-attention cache (cache) compression stages independently. A \ding{51} in the attn. column applies $\eta_\text{a}=30\%$, and a \ding{51} in the cache column applies $\eta_\text{c}=10\%$; a \ding{55} keeps that stage uncompressed.}
\label{tab:decouple_ablation}
\setlength{\tabcolsep}{8pt}
\renewcommand{\arraystretch}{1.05}
\begin{tabular}{cc|cc|cc}
\toprule
\multicolumn{2}{c|}{Stage}
& \multicolumn{2}{c}{Efficiency (batch=1)} & \multicolumn{2}{c}{Quality} \\
\cmidrule(lr){1-2} \cmidrule(lr){3-4}\cmidrule(lr){5-6}
Attn. & Cache
& TFLOPs$\downarrow$ & KV Mem.$\downarrow$ & IR$\uparrow$ & HPS$\uparrow$ \\
\midrule
\ding{55} & \ding{55} &58.50 & 7.71GB& 0.946 & 30.49 \\
\ding{51} & \ding{55} &37.96 & 7.71GB& 0.945 &30.48 \\
\ding{55} & \ding{51} &41.81 & 0.77GB& 0.948 & 30.53 \\
\rowcolor{gray!10}
\ding{51} & \ding{51} &36.94& 0.77GB&0.953&  30.50   \\
\bottomrule
\end{tabular}
\end{table}
\subsection{Hyperparameter Sensitivity Study}\label{sensi_study}
\begin{figure}[!t]
  \centering
  \includegraphics[width=\linewidth]{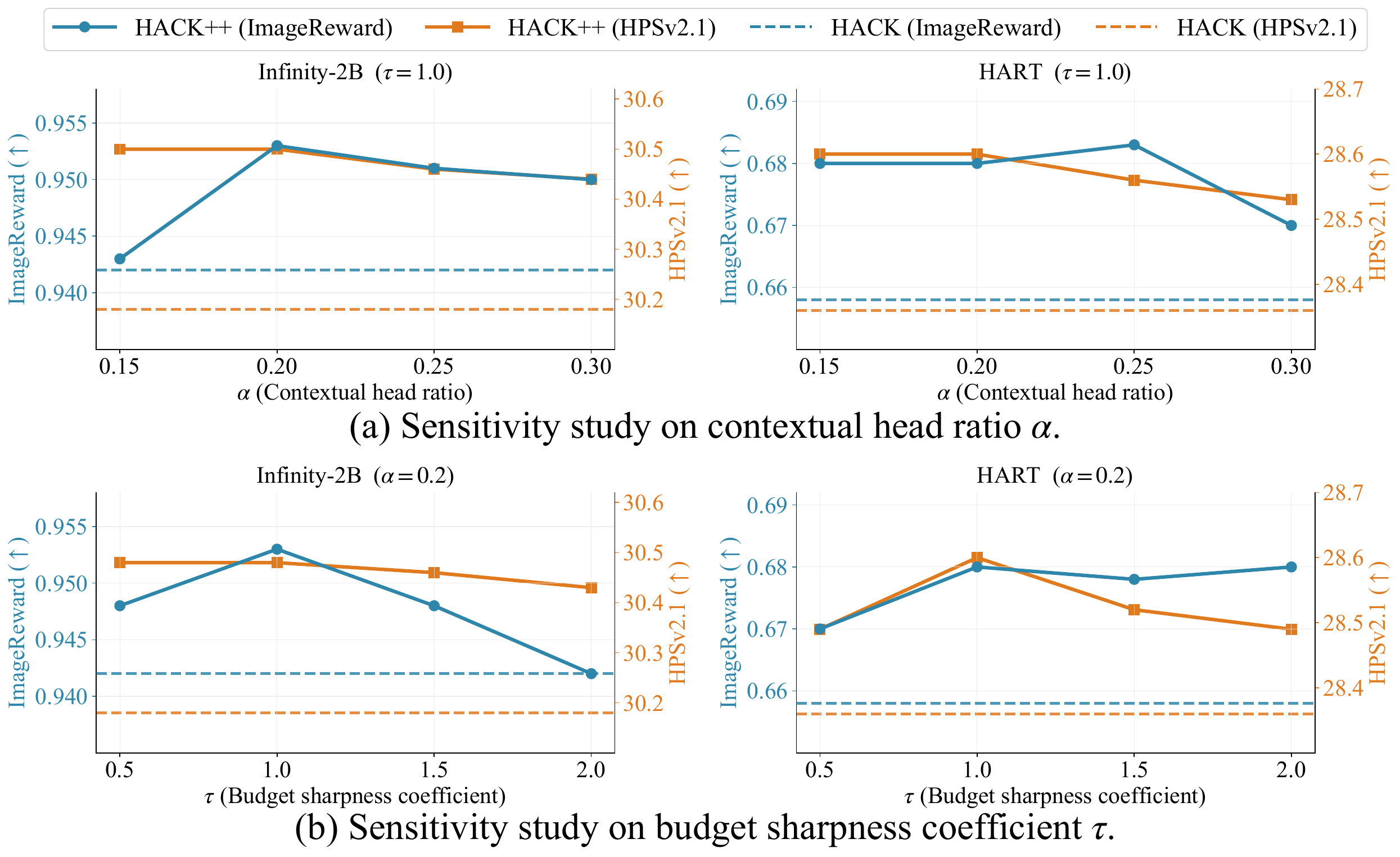}
  \caption{\textbf{Sensitivity study on hyperparameters.}
  Varying the contextual head ratio $\alpha$ (a) and budget sharpness coefficient $\tau$ (b)
  on Infinity-2B ($\eta_\text{a}=30\%,\,\eta_\text{c}=10\%$) and
  HART ($\eta_\text{a}=30\%,\,\eta_\text{c}=20\%$).
  The HACK baseline (dashed) uses $\eta_\text{a}=\eta_\text{c}=30\%$.
  HACK++ stays stable and consistently surpasses HACK under a tighter cache budget.}
  \label{fig:hyper_sensi}
\end{figure}

We study the robustness of HACK++ to its two principal hyperparameters: the
contextual head ratio $\alpha$, which controls the fraction of heads assigned the
contextual compression strategy, and the budget sharpness coefficient $\tau$, which governs
the adaptive allocation of the cache budget. We conduct two sensitivity
studies on Infinity-2B and HART:
(a) varying $\alpha$ with $\tau$ fixed, and (b) varying $\tau$ with $\alpha$ fixed,
reporting ImageReward and HPSv2.1.
As shown in Fig.~\ref{fig:hyper_sensi}, HACK++ remains stable across the entire
range of both hyperparameters and consistently surpasses the HACK baseline on both
metrics and both models, while operating under a substantially tighter cache budget.
The behavior of the two hyperparameters differs in nature: $\alpha$ is set per model
according to the attention variance distribution and stays robust within a suitable
neighborhood of this value, whereas $\tau$ remains robust across all models over a
broad band of $0.5$–$2.0$. This indicates that the effectiveness of HACK++ does not
hinge on careful hyperparameter tuning.

\begin{figure}[!t]
  \centering
  \includegraphics[width=\linewidth]{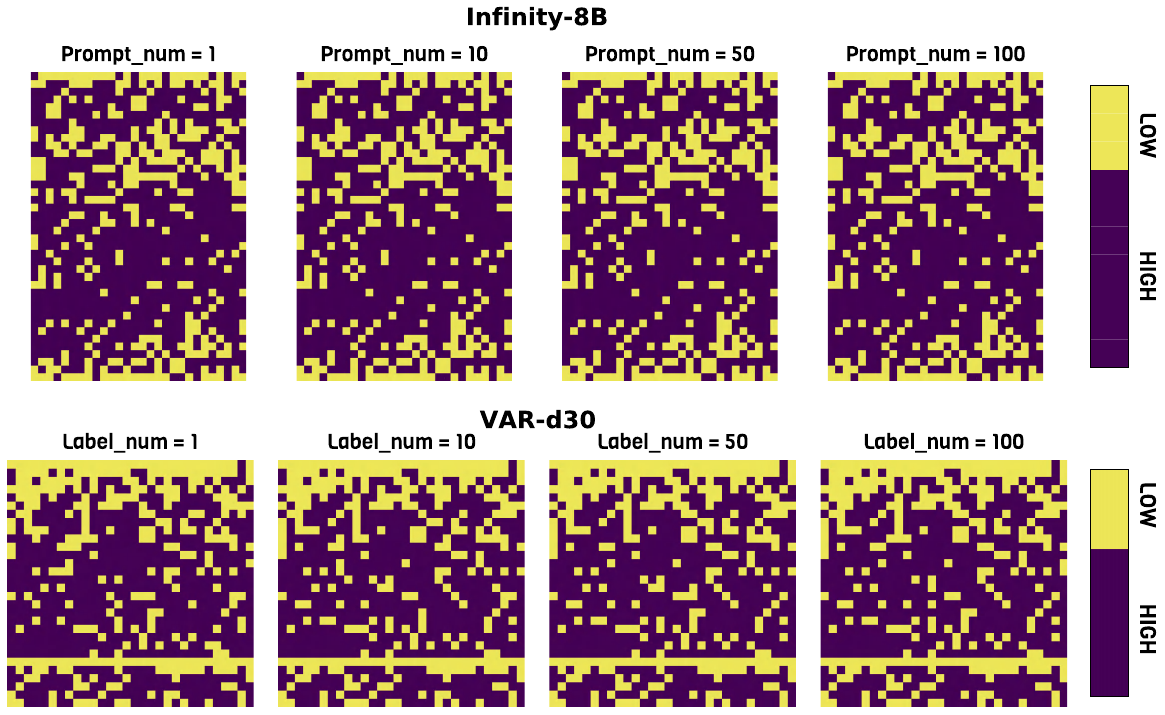}   \caption{Head classification results of Infinity-8B and VAR-d30 models with varying size of calibration set (prompts/label). Yellow indicates Contextual Heads (low variance); purple indicates Structural Heads (high variance).
  \label{fig:Head_classification}}
\end{figure}

\subsection{Impact of Calibration Set Size on Head Classification}\label{ssc}
To examine how the size of the calibration set affects the offline classification results, we vary the number of calibration samples (prompts or
class labels). As shown in Fig.~\ref{fig:Head_classification}, both Infinity-8B and
VAR-d30 yield highly stable head classifications across the entire range, from $1$
to $100$ prompts or class labels. This supports our core finding
that the functional roles of attention heads are largely invariant to the
calibration set size, reflecting a strong inductive prior in head behavior.
Consequently, a small calibration set already provides clear head-wise separability
at negligible cost.

\section{Limitations and Future Work}
Despite the effectiveness of HACK++, several directions remain. First, although attention heads in VAR models can be broadly categorized into contextual and structural types, a finer-grained taxonomy based on their varying reliance on historical scales could enable more precise importance estimation and budget allocation, further pushing the compression frontier. Second, while this work focuses on next-scale VAR generation, the head-aware perspective offers a generalizable principle that may inform efficient KV cache management across a broader class of autoregressive visual models. Finally, HACK++ instantiates this perspective in a training-free setting, yet the functional specialization of attention heads, particularly the clear separation between semantic and spatial information at early scales, opens opportunities beyond efficiency: contextual and structural heads could be selectively manipulated for controllable generation and image editing, steering semantic content or spatial layout.

\section{Conclusion}
We propose HACK++, a training-free, head-aware key-value
compression pipeline that reduces both attention complexity and
KV cache footprint for VAR models. Building on the distinct
contextual and structural attention patterns in VAR models,
HACK++ applies pattern-specific compression tailored to each
head type, and further introduces a reliance-aware cache budget
allocation to accommodate the shifting reliance on historical
scales across layers and steps. Extensive experiments on
text-to-image, class-conditional, and unified
understanding-and-generation models validate HACK++'s
effectiveness in achieving aggressive attention and KV cache
compression while maintaining generation quality, significantly
boosting inference speed and reducing memory consumption in VAR
models.

\bibliographystyle{IEEEtran}
\bibliography{ieeetran}

\end{document}